\RequirePackage{fix-cm}
\documentclass[smallextended]{svjour3}       
\smartqed  
\usepackage{graphicx}
\usepackage[table]{xcolor}
\usepackage{amsmath}
\usepackage{amssymb}
\usepackage{booktabs}
\usepackage{algorithm}
\usepackage{algorithmic}
\usepackage{pifont}
\usepackage{xcolor}
\usepackage{multirow}
\usepackage{natbib}
\usepackage{hyperref}
\usepackage{array}
\usepackage{wrapfig}
\usepackage{subcaption}
\usepackage{lipsum}
\newcommand{\rv}[1]{\textcolor{black}{#1}}
\definecolor{lblue}{RGB}{0,0,0}
\newcolumntype{C}{>{\color{black}}c}
\newcolumntype{F}{>{\color{black}\centering}m{0.65cm}}

\def\etal{\emph{et al.}}
\def\ie{\emph{i.e.}}
\begin{document}

\title{\rv{LADDER: Latent Boundary-guided Adversarial Training}
}


\author{Xiaowei~Zhou         \and
        Ivor~W.~Tsang  \and
        Jie~Yin
}


\institute{Xiaowei~Zhou \at
              Australian Artificial Intelligence Institute (AAII), School of Computer Science, FEIT, University of Technology Sydney, Ultimo, NSW 2007, Australia\\
              and \\
              Data61, CSIRO, NSW 2015, Australia\\
              \email{Xiaowei.Zhou@student.uts.edu.au}           
           \and
           Ivor~W.~Tsang \at
           Australian Artificial Intelligence Institute (AAII), School of Computer Science, FEIT, University of Technology Sydney, Ultimo, NSW 2007, Australia\\
           and \\
           Center for Frontier AI Research Agency for Science, Technology and Research (A*STAR) Singapore\\
           \email{ivor.tsang@uts.edu.au}
           \and
           Jie~Yin \at
           Discipline of Business Analytics, The University of Sydney, NSW 2006, Australia\\
           \email{jie.yin@sydney.edu.au}
}

\date{Received: date / Accepted: date}

\maketitle

\begin{abstract}
Deep Neural Networks (DNNs) have recently achieved great success in many classification tasks. Unfortunately, they are vulnerable to adversarial attacks that generate adversarial examples with a small perturbation to fool DNN models, especially in model sharing scenarios. Adversarial training is proved to be the most effective strategy that injects adversarial examples into model training to improve the robustness of DNN models against adversarial attacks. However, adversarial training based on the existing adversarial examples fails to generalize well to standard, unperturbed test data. To achieve a better trade-off between \rv{standard} accuracy and adversarial robustness, we propose a novel adversarial training framework called \textit{LAtent bounDary-guided aDvErsarial tRaining} (LADDER) that adversarially trains DNN models on latent boundary-guided adversarial examples.
As opposed to most of the existing methods that generate adversarial examples in the input space, LADDER generates a myriad of high-quality adversarial examples through adding perturbations to latent features. The perturbations are made along the normal of the decision boundary constructed by an SVM with an attention mechanism. We analyze the merits of our generated boundary-guided adversarial examples from a boundary field perspective and visualization view. 
Extensive experiments and detailed analysis on MNIST, SVHN, {CelebA, and CIFAR-10} validate the effectiveness of LADDER in achieving \rv{a better trade-off between standard accuracy and adversarial robustness as compared with vanilla DNNs and competitive baselines.}

\keywords{Adversarial training \and Adversarial robustness \and Boundary-guided generation}
\end{abstract}

\section{Introduction}
\label{intro}
In recent years, deep neural networks (DNNs) have been successfully applied \rv{to empower many advanced applications}, such as image processing~\citep{huang2017densely,shi2021generative}, speech generation~\citep{seide2011conversational,amodei2016deep} and natural language processing~\citep{fedus2018maskgan,alikaniotis2016automatic}. \rv{Nevertheless}, training a DNN model often requires large amounts of labeled data and significant efforts of parameter tuning. As such, \rv{this catalyzes new ways of developing} DNN models as a service that can be shared by a third party. This \rv{development} leads to many offline and online Machine Learning as a Service (MLaaS) platforms that provide shared services for various tasks based on DNN models, such as image and video analysis from the AWS pre-trained AI Services~\citep{Amazon}, and powerful image analysis from Google Cloud Vision~\citep{Google}. 
\vspace{-0.6cm}
\begin{figure}[htbp]
    \centering
    \includegraphics[scale=0.9]{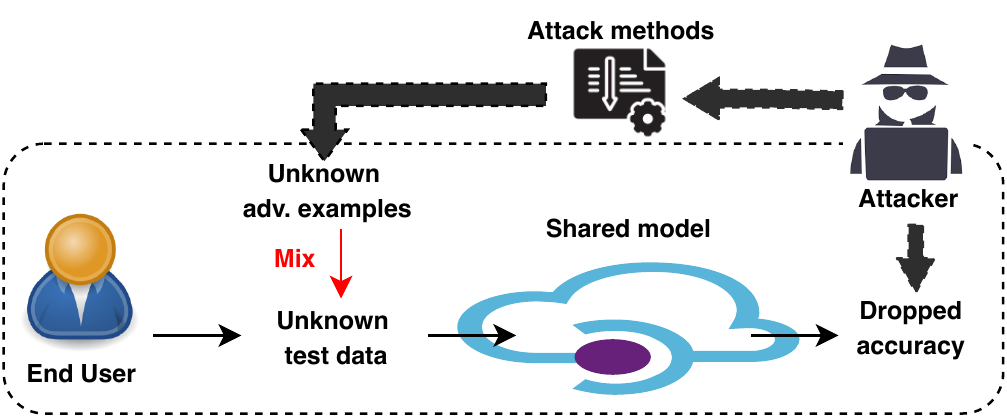}
    \caption{Attacks in model sharing scenarios. The end user performs classification tasks through the shared model without the knowledge on adversarial examples generated by attackers to attack the shared model.}
    \label{fig:model_share}
\end{figure}

In such model sharing scenarios, the increasing use of DNN models, however, has raised serious security and reliability concerns. As illustrated in Fig.~\ref{fig:model_share}, the service provider trains a DNN model using the training data that they have collected, expecting it to achieve high classification accuracy on test samples \rv{with} similar distributions. 
End-users then provide their own test samples, mostly unknown to the service provider in advance, into the shared model to obtain prediction results. However, very often, test samples are likely to be mixed with unknown adversarial examples~\citep{szegedy2013intriguing,yuan2019adversarial}, which are generated by attackers by adding small and hardly visible perturbations to standard test samples.
What makes it even worse is that adversarial examples can be generated by various types of unknown attacks, resulting in a dramatic drop in classification accuracy. This is due to the fact that the shared DNN models are not robustly trained to defend against \rv{unknown adversarial attacks} before they are released as a service. Thus, research works have been proposed to improve the robustness of DNN models \rv{against adversarial attacks.}

Adversarial training ~\citep{Goodfellow2014Exp,shaham2018understanding,shrivastava2017learning} is \rv{one of the most successful techniques for improving} the robustness of DNN models \rv{against} adversarial attacks. Its key idea is to augment the training data with adversarial examples to retrain DNN models. However, as studied in~\citep{tsipras2018robustness}, there exists a trade-off between \rv{standard accuracy on clean data and adversarial robustness}. Better robustness often leads to worse classification accuracy on clean data. To address this, several recent studies~\citep{zhang2019theoretically,Ding2020MMA,wang2020improving} have been proposed to add additional regularization terms into the loss function of adversarial training. {These methods employ the modified adversarial training loss function with regularization to achieve better generalization and adversarial robustness.} This has raised a research question: Could adversarial examples generated by the existing methods account for the trade-off between standard accuracy and adversarial robustness? This motivates us to explore the curses and blessings of adversarial examples for achieving a better trade-off between standard accuracy and robustness \rv{against adversarial attacks.} 

The adversarial examples generated by the existing methods suffer from two major curses. \rv{First, most of these methods~\citep{Goodfellow2014Exp,madry2017towards,papernot2016limitations,carlini2017towards} generate adversarial examples through adding a small perturbation to legitimate samples in the input space}, which often only craft adversarial examples of repeating patterns. The DNN models adversarially trained with these examples would be effective only in defending very specific types of adversarial attacks, but still vulnerable to other adversarial attacks. Thus, this presses a need \rv{for increasing} the diversity of generated adversarial examples so that DNN models can fully explore the unknown data space to improve \rv{the robustness against adversarial attacks.} 
\begin{figure}[tbp]
    \centering
    \includegraphics[width=0.98\linewidth]{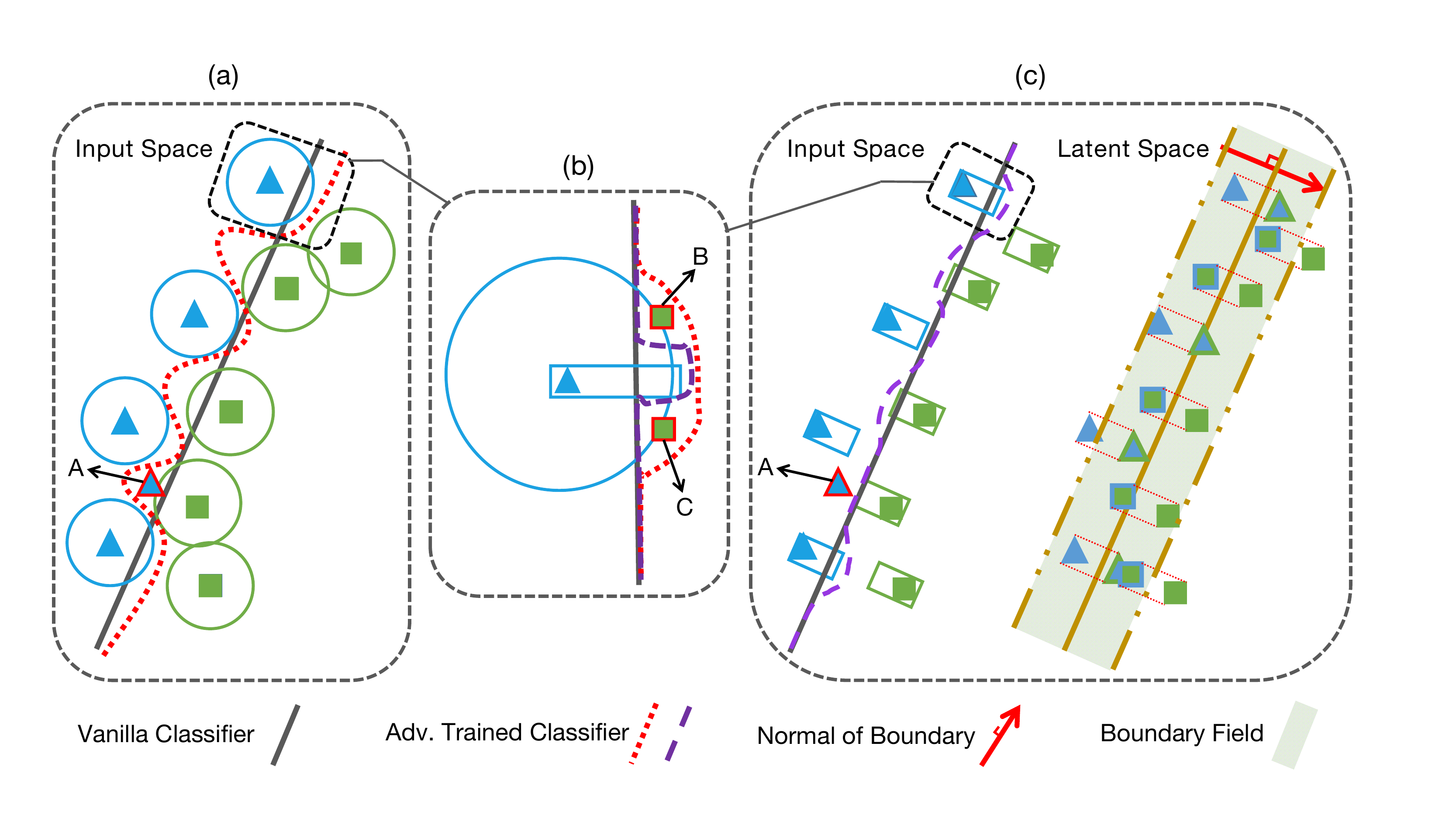}
    \caption{(a): Adversarial examples generated by the existing methods are often within a ball in the input space, which leads the decision boundary of the adversarially trained classifier (dotted line) to dramatically change. The legitimate sample, sample A, would be misclassified. (c): Our LADDER method perturbs latent features of samples within the boundary field in the latent space. The generated adversarial examples would reside in a more restricted area. Sample A would thus be classified correctly. (b): The adversarially trained classifier by existing adversarial examples (red dotted line) would misclassify legitimate samples (sample B and C), which would hurt the standard accuracy. The adversarially trained classifier using our method (purple dashed line) would change less remarkably than the existing one (red dotted line), which would correctly classify the samples.}
    \label{fig:motivation}
\end{figure}

Second, the generated adversarial samples might hurt the standard accuracy of DNN models on clean (legitimate) samples. As shown in Fig.~\ref{fig:motivation} (a), the adversarial examples generated by most existing methods are within a ball in the input space. When adversarially trained with these samples, the decision boundary of the DNN classifier would dramatically change as compared with the original one. This leads legitimate samples (sample A) to be misclassified. Moreover, as perturbations are added in the input space, the generated adversarial examples would contain lots of noise. This can be seen from adversarial examples (in Fig.~\ref{fig:our_AdSm}) generated by FGSM~\citep{Goodfellow2014Exp} or JSMA~\citep{papernot2016limitations}. Consequently, it would hurt the standard accuracy of adversarially trained models. 
This inspires us to generate adversarial examples of better quality through the latent space. 

Furthermore, existing methods treat all examples \rv{equally} while generating adversarial examples, but neglect the decision boundary information. Our work stems from the observation that incorporating information about the decision boundary in the latent space into the generation of adversarial examples would be a blessing to achieve better generalization on standard clean data. As shown in Fig.~\ref{fig:motivation}(c), latent features of samples are perturbed along the normal of decision boundary to move within the boundary field, \ie, the nearby area of the decision boundary. The generated adversarial examples would thus reside in a restricted area and prevent the adversarially trained classifier from misclassifying legitimate samples (sample B and C), which, however, could be misclassified by existing methods. 
Motivated by this observation, we focus on how to leverage the decision boundary to guide the generation of adversarial examples, aiming to achieve a better trade-off between standard accuracy and adversarial robustness.

In this paper, we propose a novel adversarial training framework called Latent Boundary-guided Adversarial Training (LADDER), which adversarially trains DNN models with myriad adversarial examples generated based on the decision boundary in a latent space.
\rv{Unlike some of the existing methods that operate in the input space and generate noisy adversarial examples of repeated patterns, LADDER generates high-quality and diverse adversarial examples by adding perturbations to latent features.} The generation of adversarial examples is guided by the normal of decision boundary in the latent space, which is learned via a linear support vector machine (SVM)~\citep{boser1992training} with an attention mechanism. After adversarial training on generated adversarial examples, the adversarially trained DNN model \rv{is effective in achieving a trade-off between standard accuracy and robustness against adversarial attacks.} Comprehensive experiments and analyses are conducted on MNIST, SVHN, CelebA, and CIFAR-10 to verify the effectiveness of the proposed method. 

\rv{The novelty and contribution of this paper are three-fold:}
\begin{itemize}
    \item We analyze the curses and blessings of adversarial examples for adversarial training and explain the advantages of latent boundary-guided solution.
    \rv{\item We propose a new method called LADDER that generates high-quality and diverse adversarial examples by adding boundary-guided perturbations in a latent feature space.}
    \item After adversarial training on the generated adversarial examples, LADDER achieves a better trade-off between standard accuracy and adversarial robustness as compared with vanilla DNNs and competitive baselines.
\end{itemize}

\section{Related Work}
This section reviews two branches of related literature: adversarial attack methods and adversarial defence methods.

\subsection{Adversarial Attack}

{From the methodology point of view, most of the existing adversarial attack methods can be grouped into two categories: \rv{\textit{gradient-based attacks}~\citep{Goodfellow2014Exp,madry2017towards,papernot2016limitations} and \textit{decision-based attacks}~\citep{carlini2017towards,song2018constructing,moosavi2016deepfool,su2019one}}}.

\paragraph{\textbf{Gradient-based Attacks.}}
Gradient-based \rv{attacks mainly add perturbations} in the direction of the gradient of loss function with respect to the input sample. Goodfellow \etal~\citeyearpar{Goodfellow2014Exp} proposed the fast gradient sign method (FGSM) that uses the sign of gradient ($\nabla_{\mathbf{x}}J(\theta, \mathbf{x}, y)$) of loss function with respect to input examples as perturbation. 
Built upon FGSM, the one-step attack method, Madry \etal~\citeyearpar{madry2017towards} proposed a multi-step attack method called PGD. PGD iteratively uses the gradient information and generates adversarial examples on the results of the last step. \rv{Similarly, DI$^2$-FGSM~\citep{xie2019improving} generates adversarial examples on the results of last step, but it uses the gradients of stochastic transformed inputs rather than the original ones.} Papernot \etal~\citeyearpar{papernot2016limitations} introduced saliency map based on Jacobian matrix into the generation of adversarial examples. The saliency values computed by forward derivative of a target model are used as an indicator to determine the locations in the input examples to add perturbations. This method is called Jacobian saliency map attack (JSMA). 

\paragraph{\textbf{Decision-based Attacks.}}
Decision-based \rv{attacks} manipulate the labels of training data to make the learned DNN model beneficial to their specific purposes. This line of methods uses Eq.~(\ref{eqa:adv}) as a measure, \ie, changing the original label to target label, to generate adversarial examples. For a given classifier $f$, the predicted label of an input sample $\mathbf{x}$ is defined as $f(\mathbf{x})$, and $f(\mathbf{x}')$ is the label of adversarial example $\mathbf{x}'$: 
\begin{equation}
\begin{aligned}
    \mathbf{x}' = \mathbf{x} + \boldsymbol{\delta}, \mbox{ s.t. }
    f(\mathbf{x}') \neq f(\mathbf{x}),  
    \label{eqa:adv}
\end{aligned}
\end{equation}
where $\boldsymbol{\delta}$ is a small perturbation added to an input sample $\mathbf{x}$. $\mathbf{x}'$ is the generated adversarial example. Carlini and Wagner~\citeyearpar{carlini2017towards} proposed an approach, CW, to generate adversarial examples by adding small changes to the original images in the input space. CW \rv{tries} to minimize the distance between benign examples and adversarial ones, while enforcing the labels of adversarial examples as the targeted ones. A deep learning based attack method was developed by~\citep{song2018constructing}. This method explored the AC-GAN~\citep{odena2017conditional} latent space to generate adversarial images, which could most likely mislead the targeted classifier. 
\rv{GeoDA~\citep{rahmati2020geoda} estimates the decision boundary for each data point in the input space to generate adversarial examples. In contrast, our LADDER trains an SVM to obtain the decision boundary between any two classes in the latent space.}

\paragraph{\textbf{Other Adversarial Attacks.}}
\rv{Recently, distribution-based methods have also been proposed for adversarial attack. DAA~\citep{zheng2019distributionally}, HMCAM~\citep{wang2020hamiltonian} and $\mathcal{N}$attack~\citep{li2019nattack} generate diverse adversarial examples through modeling their probability distribution in the input space. The goal of DAA~\citep{zheng2019distributionally} is to generate globally optimal adversarial examples and $\mathcal{N}$attack~\citep{li2019nattack} aims to develop a powerful black-box adversarial attack method. HMCAM~\citep{wang2020hamiltonian} generates a sequence of adversarial examples to improve adversarial robustness. In contrast, our LADDER, as one adversarial training based defence method, generates adversarial examples individually for each input to achieve a better trade-off between standard accuracy and adversarial robustness. }

\rv{Apart from distribution-based methods, Croce \etal~\citeyearpar{croce2020reliable} proposed an ensemble attack method called AutoAttack. This method firstly extends the PGD attack method into APGD$_{CE}$ by automatically choosing step sizes and APGD$_{DLS}$ by using a difference of logits ratio (DLR) loss. Then, four attack methods, including APGD$_{CE}$, APGD$_{DLS}$, FAB~\citep{croce2020minimally} and square attack~\citep{andriushchenko2020square} are combined as AutoAttack.}

{From the knowledge accessibility point of view, the adversarial attacks can be divided into \textit{white-box attacks} and \textit{black-box attacks}. Under white-box attack settings, the attackers have access to full knowledge about the target model, \ie, model structure and model parameters. On the contrary, the attackers have no knowledge about the target model under black-box attack settings. In this work, we are mainly concerned \rv{with} model sharing scenarios, where model structure and model parameters are unknown to attackers. Thus, we focus on defending black-box attacks.}

\subsection{Adversarial Defence} 
For various types of adversarial attacks, a key research question is, how can we improve the adversarial robustness of a DNN model before it is deployed as a service? In response, adversarial defence strategies have been proposed to mitigate the effect of adversarial attacks~\citep{papernot2016distillation,papernot2017extending,samangouei2018defense, meng2017magnet, guo2018countering, kannan2018adversarial, mustafa2019adversarial, Xiao2020Enhancing}. 

\paragraph{\textbf{Gradient Masking}} \rv{Gradient masking methods ~\citep{papernot2017practical} construct a model which does not provide useful gradients to be attacked. For example,}
\textit{Defensive distillation}~\citep{papernot2016distillation,papernot2017extending} learns a smooth targeted defence model with $d$ training times, where the predicted labels from $(d-1)$-th model are used as ground truth to train the $d$-th model except for the first time. 

\paragraph{\textbf{Clean Sample Reconstruction}}
{Defense-GAN}~\citep{samangouei2018defense}\rv{, DIPDefend~\citep{dai2020dipdefend} and} MagNet~\citep{meng2017magnet} are \rv{three} typical methods that 
remove the perturbation added on adversarial examples to \rv{reconstruct a clean sample similar to the legitimate one}. The \rv{reconstructed} examples can be easily \rv{recognised} by the model, compared with adversarial examples. 

\paragraph{\textbf{Adversarial Training.}}
Among others, \textit{adversarial training}~\citep{Goodfellow2014Exp,shaham2018understanding,shrivastava2017learning,kurakin2016adversarial} is proved as one of the most effective defence methods, which augments the training data with adversarial examples when training the targeted model. It can be achieved either by training the targeted model with original samples augmented with adversarial examples~\citep{kurakin2016adversarial} or with a modified loss function~\citep{Goodfellow2014Exp}. \rv{Our LADDER falls into the realm of adversarial training based defence methods.}

Recently, Tsipras \etal~\citeyearpar{tsipras2018robustness} found the adversarial robustness is at odds with the standard accuracy on clean samples. Several recent methods were proposed to trade the adversarial robustness off the standard accuracy. \rv{Most of these methods considered adding different regularization terms to the adversarial training loss to achieve a better trade-off~\citep{zhang2019theoretically,Ding2020MMA,wang2020improving, zhang2021geometry}}. \rv{TRADES~\citep{zhang2019theoretically} is a regularization-based method} that minimizes the loss between the predicted labels and ground truth of legitimate samples as well as the 'difference' between the \rv{predictions} of legitimate samples and the corresponding adversarial examples. The 'difference' term was regarded as the regularization. 
\rv{GAIRAT~\citep{zhang2021geometry} uses the distance to the decision boundary to assign a weight for each adversarial example in the adversarial training loss. The weight is estimated based on the difficulty of attacking the input by PGD rather than the actual distance to the decision boundary. On the contrary, our LADDER derives an explicit decision boundary to generate adversarial examples for adversarial training.} It is worth noting that, the existing regularization based methods and our LADDER method provide two different views to achieve a better trade-off between the standard accuracy and adversarial robustness. Our LADDER method can be used in combination with these regularization based methods to further boost their performance (See our empirical results in Section~\ref{exp:CompWithTrades}).

\rv{Different from the existing methods that use a regularized loss to improve the trade-off, AVmixup~\citep{lee2020adversarial} is a data augmentation method that uses linear interpolation to acquire the augmented examples in the input space based on adversarial examples generated by PGD attack. However, as proved in Manifold Mixup~\citep{verma2019manifold}, AVmixup may produce less semantically meaningful examples as a result of linear interpolation in the input space. In contrast, our LADDER method alleviates this issue by adding perturbations along the normal of the decision boundary in the latent space.} 

\section{Latent Boundary-guided Adversarial Training}


To achieve a better trade-off between standard accuracy and adversarial robustness, LADDER aims to generate better adversarial examples based on the decision boundary constructed in a latent space. Perturbations are added to latent features along the normal of decision boundary and inverted to the input space by the trained generator. 
Through adversarial training with the generated adversarial examples, LADDER achieves a better trade-off between standard accuracy and adversarial robustness. 

For clarity, we first define key notations and symbols. Define $\mathbb{X} = \{\mathbf{x_1},...,\mathbf{x}_n\}$ as the set of samples, where $n$ is the number of samples. For each sample $\mathbf{x}_i$, $\mathbf{z}_i$ is the latent feature vector extracted from a trained DNN model. $\mathbf{d}$ is the normal of \rv{the} decision boundary. $\epsilon$ is the perturbation added to the latent feature vector $\mathbf{z}_i$. $\mathbf{\hat{x}}_i$ is the generated sample from the latent feature $\mathbf{z}_i$. 



\subsection{Latent Boundary-guided Generation}
\label{sec:attack}

The adversarial examples are generated through perturbating the latent space, guided by the decision boundary, which is obtained from an attention SVM learned on latent features. 





\subsubsection{Boundary-guided Attention} {To approximate local decision boundaries in the latent space, we train a linear SVM with an attention mechanism~\citep{zhang2019self} with latent features of a DNN model. This idea is grounded on the theoretical proof by ~\cite{li2018decision} that, the last layer of neural networks trained by cross-entropy loss converges to a linear SVM. For any neural network used for binary or multi-class classification, when the cross-entropy loss gradually approaches to $0$, the last layer weights of neural networks would converge to the solution of an SVM. Specifically, we use the latent feature $\mathbf{z}$, \rv{the input to the last layer} of the DNN model to train a linear SVM with an attention mechanism. The trained linear SVM provides an explicit margin as compared to other linear models such as a linear mixture model.}

\begin{figure}[t]
    \centering
    \includegraphics[width=1.0\linewidth]{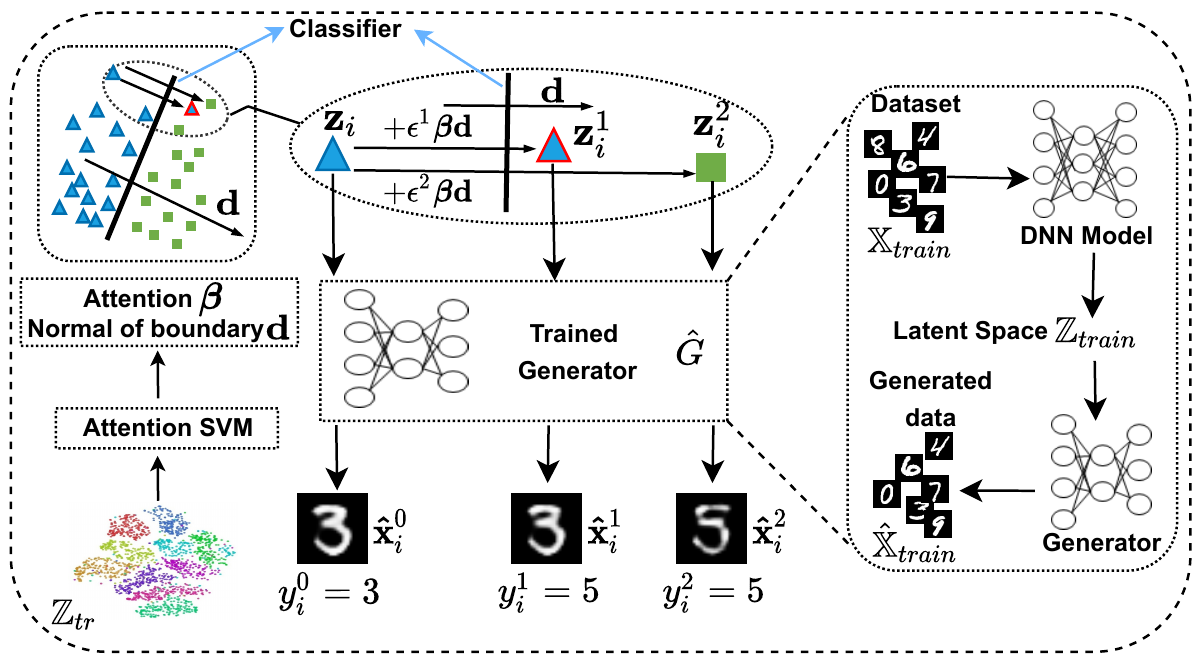}
    \caption{Overview of \textit{Latent Boundary-guided Adversarial Training}. LADDER generates adversarial examples by perturbing latent features alongside the normal of decision boundary obtained from an SVM with an attention mechanism. These generated adversarial examples are inverted to the input space via a trained generator to adversarially train the DNN model. $\mathbf{z}_i$ is the latent feature of one original sample $\mathbf{x}_i$; $\boldsymbol{\beta}$ is attention weight; $\mathbf{d}$ is the normal of \rv{the} decision boundary; $\mathbf{z}_i^1$ and $\mathbf{z}_i^2$ are the perturbed latent features for generation; $\mathbf{\hat{x}}_i^j\ (j=0,1,2)$  are the generated images after perturbing the latent features; $y_i^j\ (j=0,1,2)$ are the predicted labels of the generated images.}
    \label{fig:flowchat}
\end{figure}


By employing an attention mechanism when training the linear SVM, our aim is to capture a better representation with different weights assigned to different elements of latent features. 
To this end, an attention layer is added to process latent features, before passing them to the SVM. 
The attention layer is defined as follows:
\begin{equation}
\boldsymbol{\alpha}_i= \tanh\left(conv(\mathbf{z}_i)\right)\!, \, \boldsymbol{\beta}_i^j \!=\!\! \frac{\exp(\boldsymbol{\alpha}_i^j)}{\sum_{j=1}^N \exp(\boldsymbol{\alpha}_i^j)}\!, \mbox{and}\  \mathbf{z}_i^{att} \!=\! \boldsymbol{\beta}_i  \mathbf{z_i},
\end{equation}
{where $\mathbf{z}_i$ is a latent feature vector, \rv{the input to the last layer} of the DNN model}; $conv$ is the convolutional operation; $\tanh$ is the activation function; $\boldsymbol{\beta}_i$ is the attention weight vector that can be learned; $\mathbf{z}_i^{att}$ is the output after applying attention weights on latent feature vector $\mathbf{z}_i$. 

\subsubsection{Latent Feature Perturbation} After training an SVM with attention, the latent features of each sample are perturbed along the normal of the decision boundary of the SVM. The normal $\mathbf{d}$ provides a direction to guide the generation, where the attention weight $\boldsymbol{\beta}$ captures the importance of different components of latent features to move across the boundary. Different perturbations $\epsilon$ can be added to the same latent features $\mathbf{z}_i$ to obtain the perturbed latent features $\mathbf{z}_i^j$ by: 
\begin{equation}
        \mathbf{z}_i^j = \mathbf{z}_i + \epsilon^j \boldsymbol{\beta}_i \mathbf{d},
    \label{eqa:modifyFea}
\end{equation}
where vector $\mathbf{d}$ is the normal of the decision boundary of the linear SVM; $\boldsymbol{\beta}_i$ is the attention weight vector obtained by the SVM for each sample; $\epsilon^j>0$ represents the perturbation; $j$ is the index of different perturbation.

When the perturbation is big enough, the class label of the perturbed latent features would change from positive to negative, or vice versa. That means it would cross the decision boundary of the DNN model. As shown in Fig.~\ref{fig:flowchat}, perturbed latent features $\mathbf{z}_i^1$ move from the left side of the decision boundary to the right side. As the perturbation continues to increase, perturbed latent features $\mathbf{z}_i^2$ would move far away from the decision boundary. The effect of perturbation will later be empirically investigated in Section~\ref{exp:perturbation}.

\subsubsection{Boundary-guided Generation} 

To enable humans to understand what changes happen in the input space, caused by the perturbations to latent features, we train a generator to invert the perturbed latent features to the input space.  



For a specific DNN model (\ie, LeNet), we learn a generator $\hat{G}$ on the training set $\mathbb{X}_{train}=\{\mathbf{x}_1,...,\mathbf{x}_n\}$ to map latent features to the input space. As shown in Fig.~\ref{fig:flowchat}, each sample in the training set is fed into the DNN model, to extract the corresponding latent features. The output of the last fully connected layer of a DNN model can be used to construct the set of latent features $\mathbb{Z}_{train}=\{\mathbf{z}_1,...,\mathbf{z}_n\}$. Sample $\mathbf{x}_i$ and its corresponding latent features $\mathbf{z}_i$ are fed to the designed generator. 
The objective function of $G$ over the neural network class $\mathcal{G}$ is defined as follows:
\begin{equation}
    \hat G =\arg \min\limits_{G\in\mathcal{G}} n^{-1} \sum_{i=1}^{n}\left\|\mathbf{x}_{i} - G(\mathbf{z}_i)\right\|_p^p,
    \label{eqa:gene}
\end{equation}
where $\|\cdot\|_p$ denotes $L_p$ norm, $p=1~or~p = 2$ in this paper; $\mathbf{z}_i=\Phi(\mathbf{x}_{i})$, where $\Phi$ is a feature extractor in a DNN model.
A mapping between the latent space and the input space is learned by optimizing Eq. (\ref{eqa:gene}). 

The reconstructed sample $\hat{\mathbf{x}}_i$ can be obtained by passing $\mathbf{z}_i$ to the trained generator $\hat{G}$, that is, $\hat{\mathbf{x}}_i = \hat G\left(\mathbf{z}_{i}+\epsilon  \boldsymbol{\beta} \mathbf{d}\right)$. For example, in Fig.~\ref{fig:flowchat}, different latent features ($\mathbf{z}_i$, $\mathbf{z}_i^1$ and $\mathbf{z}_i^2$) are fed into the trained generator $\hat{G}$ to obtain the corresponding samples $\hat {\mathbf{x}}_i^j\ (j=0,1,2)$ in the input space. 

When the generated samples are fed into the targeted DNN model, the predicted labels should be the same as the ground-truth label $\hat{y}$. That is, the following equation should be satisfied:
\begin{equation}
     f\left(\hat{G}(\mathbf{z}_i^j)\right) = 
     \left\{
        \begin{aligned}
        \hat{y}_i,\ if\ \mathbf{z}_i^j\ not\ cross\ the\ boundary\\
        \hat{y}_{other},\ if\ \mathbf{z}_i^j\ cross\ the\ boundary
        \end{aligned}
    \right.
    \label{eqa:label_equal}
\end{equation}
As shown in Fig.~\ref{fig:flowchat}, samples $\mathbf{\hat{x}}_i^0$ and $\mathbf{\hat{x}}_i^2$ are generated from $\mathbf{z}_i$ and $\mathbf{z}_i^2$, with $0$ and $\epsilon^2$ perturbation added, respectively. Among them, $\mathbf{z}_i$ does not cross the boundary and the predicted label of $\mathbf{\hat{x}}_i^0$ is still the ground-truth label, $y_i^0=3$. For $\mathbf{z}_i^2$ that has crossed the boundary, the predicted label of its corresponding sample $\mathbf{\hat{x}}_i^2$ has changed to 5, $y_i^2=5$. This satisfies the rules specified by Eq.~(\ref{eqa:label_equal}).

However, Eq.~(\ref{eqa:label_equal}) does not always hold for some perturbed latent features. Fig.~\ref{fig:flowchat} provides such an illustration. 
The perturbed $\mathbf{z}_i^1$ has crossed the boundary and the predicted label $y_i^1$ of its generated sample $\mathbf{\hat{x}}_i^1$ has also changed to $5$. However, the ground-truth label of $\mathbf{\hat{x}}_i^1$ is still $3$.
Such samples, whose predicted labels of the reconstructed samples and ground-truth labels are inconsistent, are adversarial examples that are effective to attack the targeted DNN model.

\subsection{Latent Boundary-guided Adversarial Training}

\rv{To improve the adversarial robustness of the DNN models, we adopt the adversarial training method~\citep{Goodfellow2014Exp,shaham2018understanding} that uses the generated adversarial examples to augment the training data for retraining. 
Specifically, we get the perturbed latent features $\mathbf{z}_i^j$ through Eq.~(\ref{eqa:modifyFea}) and pass $\mathbf{z}_i^j$ to the trained generator $\hat{G}$ to generate adversarial examples. Then, we adversarially train the DNN model through the following adversarial loss function:} 
\begin{equation}
    \Tilde{J} = \alpha J\left(\boldsymbol{\theta}; \mathbf{x}, y\right) + (1-\alpha)J\left(\boldsymbol{\theta}; \hat{G}\left(\mathbf{z}+\epsilon \boldsymbol{\beta} \mathbf{d}\right),y\right),
    \label{eqa:loss}
\end{equation}
where $J(\boldsymbol{\theta})$ is the original loss function of the DNN model and $\boldsymbol{\theta}$ is the parameter of the targeted DNN model. The first and second term is the loss for the original training samples $\mathbf{x}$ and the generated adversarial examples $\hat{G}\left(\mathbf{z}+\epsilon \boldsymbol{\beta} \mathbf{d}\right)$, respectively. $\alpha$ is the weighting factor that trades off the two terms, which is usually set as 0.5. 
Through adversarial training on the generated boundary-guided adversarial examples, the \rv{adversarially trained} DNN model can achieve a better trade-off between standard accuracy and adversarial robustness. \rv{Without loss of generality, other consistency losses~\citep{zhang2018mixup,liu2021certainty,verma2019interpolation} in semi-supervised learning can also be used here, but they require additional modifications to be adapted for our adversarial training purposes. }

\textbf{Complexity Analysis.} Compared with the adversarial training based defence methods that generate adversarial examples in the input space, the extra overhead of LADDER mainly lies in the construction of a linear SVM and the training of our generator. The complexity of training a linear SVM is $\mathcal{O}(n^2)$, where $n=400$ is the number of samples used to train the SVM in our method. The complexity of training our generator is related to the number of layers and number of weights in the generator. After our generator is trained, the generation of adversarial examples is just one forward propagation of the trained generator. For the adversarial training part, our method has the same computational complexity as we use the original adversarial training loss function to adversarially train the model.

\section{Experimental Evaluation}
\label{sec:exp}

In this section, we present experimental results to show the effectiveness of our method in achieving a better trade-off between standard accuracy and adversarial robustness. We conduct extensive experiments on MNIST~\citep{LecunMnist}, SVHN~\citep{netzer2011reading}, { CelebA~\citep{liu2015faceattributes}, and CIFAR-10~\citep{krizhevsky2009learning}} from \rv{four} perspectives. The source code of our implementations is provided\footnote{\url{https://github.com/zhouxiaowei1120/LADDER}}.

\textbf{P1: Blessings of Adversarial Examples.} To show the merits of our latent boundary-guided adversarial examples, we visualize and analyse the generated adversarial examples. (Section~\ref{exp:blessing}).

\textbf{P2: Standard Accuracy \rv{and Adversarial Robustness.}} \rv{We evaluate the standard accuracy and adversarial robustness of different adversarially trained models and demonstrate the competitiveness of our LADDER method (Section~\ref{exp:cleanAndRobust}). In model sharing scenarios, we focus on adversarial robustness against black-box attacks.} Detailed experiments on adversarial robustness against white-box attacks can be found in \rv{Appendix~\ref{robustWhiteBox}}).

\textbf{P\rv{3}: Effect of Perturbation.} We investigate how perturbation impacts the performance of our LADDER method. (Section~\ref{exp:perturbation})

\textbf{P\rv{4: Complement to Regularization-based Adversarial Training Methods.}} \rv{We verify the complement effect of LADDER to the existing regularization-based adversarial training methods to achieve a better trade-off between standard accuracy and adversarial robustness. (Section~\ref{exp:CompWithTrades})}



\subsection{Experiments Settings}
\label{expSet}

\paragraph{\textbf{Datasets and Shared DNN Models.}} 
We conduct our experiments on {four} datasets: MNIST (one grey digits dataset), SVHN (one colorful digits dataset), {CelebA (one human face image dataset), and CIFAR-10 (one natual image dataset)}. On the {four} datasets, {w}e use DNN models with different architectures and depths, LeNet~\citep{lecun1995learning}, SVHNNet (shallow VGG model){, CelebANet (deep VGG model~\citep{simonyan2014very}), and CifarNet (ResNet18)} as the targeted classifiers for defence in model sharing scenarios, respectively. 
Note that, on CelebA, because the size of original images is 178$\times$218, we first pre-process the images to 128$\times$128 using DLIB~\citep{dlib}. We detect faces in images and crop them into square {sizes}. Our task is the classification of smile or non-smile for an input image.


\paragraph{\textbf{Baseline Methods.}} The competing methods used for comparison are summarized as follows. FGSM~\citep{Goodfellow2014Exp}, JSMA~\citep{papernot2016limitations}, PGD~\citep{madry2017towards}, CW~\citep{carlini2017towards} and \rv{AutoAttack~\citep{croce2020reliable} are five baselines} that generate adversarial examples by adding perturbations in the input space. Song \etal~\citeyearpar{song2018constructing} is one baseline method that generates adversarial examples in the latent space.  TRADES~\citep{zhang2019theoretically} is one representative method that adds regularisation into the adversarial training loss to improve the trade-off between standard accuracy and adversarial robustness. It uses adversarial examples generated by FGSM for adversarial training. 
We also compare with another baseline called TRADES+LADDER that combines TRADES with LADDER. This baseline is used to assess whether the methods that regularize the adversarial training loss can be complemented when using adversarial examples generated by our LADDER method.

For ablation study, we compare with two variants of our LADDER method: LADDER\_cavRandom and LADDER\_Random, which use different strategies for generating adversarial examples $\hat {\mathbf{x}}_i$. LADDER\_cavRandom adds some random noise $\boldsymbol{\delta}$ on the normal of decision boundary obtained from the SVM: $\hat {\mathbf{x}}_i = \hat G\left(\mathbf{z}_{i}+\epsilon  \boldsymbol{\beta} (\mathbf{d} + \boldsymbol{\delta})\right)$. LADDER\_Random uses a random noise $\boldsymbol{\gamma}$ to replace the normal of decision boundary for generation: $\hat {\mathbf{x}}_i = \hat G\left(\mathbf{z}_{i}+\epsilon  \boldsymbol{\beta} \boldsymbol{\gamma} \right)$. The two baselines are used to show LADDER's effectiveness in using the normal of decision boundary to guide the generation of adversarial examples.  

For FGSM, PGD, JSMA and CW, we generate adversarial examples using the open-source attack library \textit{cleverhans}~\citep{papernot2016technical}. For the method of~\citep{song2018constructing}\rv{, AutoAttack~\citep{croce2020reliable}} and TRADES~\citep{zhang2019theoretically}, we use the source code released by the authors. The number of generated adversarial examples for adversarial training {on each dataset is: 4,500 on MNIST; 4,500 on SVHN; 2,000 on CelebA; and 50,000 on CIFAR-10.} \rv{The hyper-parameters used for all methods in adversarial training are summarized in Table~\ref{tab:hyper} in Appendix \ref{hyperpara}}.



\subsection{Blessings of Adversarial Examples}
\label{exp:blessing}

\subsubsection{Fidelity of Generator} We first validate the performance of our trained generator in terms of the quality of the generated samples. Here, MNIST is used as a case study to visualize and analyze the results. We train a generator on MNIST using latent features with a dimension of $500$, \rv{the input to the last layer of LeNet}. All components of the generator architecture except for activation functions are provided in Table~\ref{app:mnist_g} \rv{in Appendix~\ref{generatorArch}}. After the last convolutional layer, a sigmoid activation function is added and the loss function used is mean squared error (MSE): 
$\ell(\mathbf{x}, \hat{\mathbf{x}}) = \frac{1}{n}\sum_{i=1}^n \left( \mathbf{x}_i - \hat{\mathbf{x}}_i \right)^2$.

\begin{figure}[htbp]
    \centering
    \includegraphics[width=0.9\columnwidth]{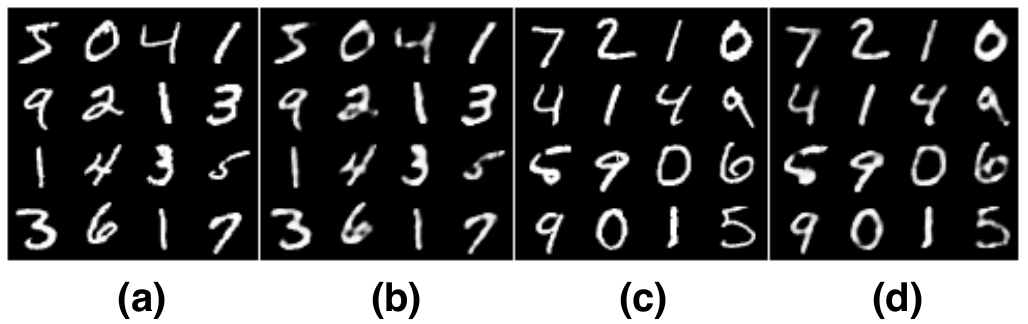}
    \caption{Reconstructed images of our generator trained on MNIST. (a) and (c) indicate the original training and test images, whereas (b) and (d) show the generated training and test images.}
    \label{fig:gene_img}
\end{figure}

We evaluate our generator through both quantitative and qualitative results. The training loss on the training dataset after \rv{1,000 epoches} and the test loss over test dataset are $0.00757$ and $0.00765$, respectively. Fig.~\ref{fig:gene_img} shows examples of reconstructed images using our generator trained on MNIST, where (a) and (c) are the original training and test images, while (b) and (d) are the generated training and test images. The generated images are very similar to the original ones. This indicates that the trained generator is able to capture the mapping between the latent space and the input space.

\begin{figure*}
\begin{subfigure}{0.47\columnwidth}
    \centering
    \includegraphics[width=1.0\textwidth]{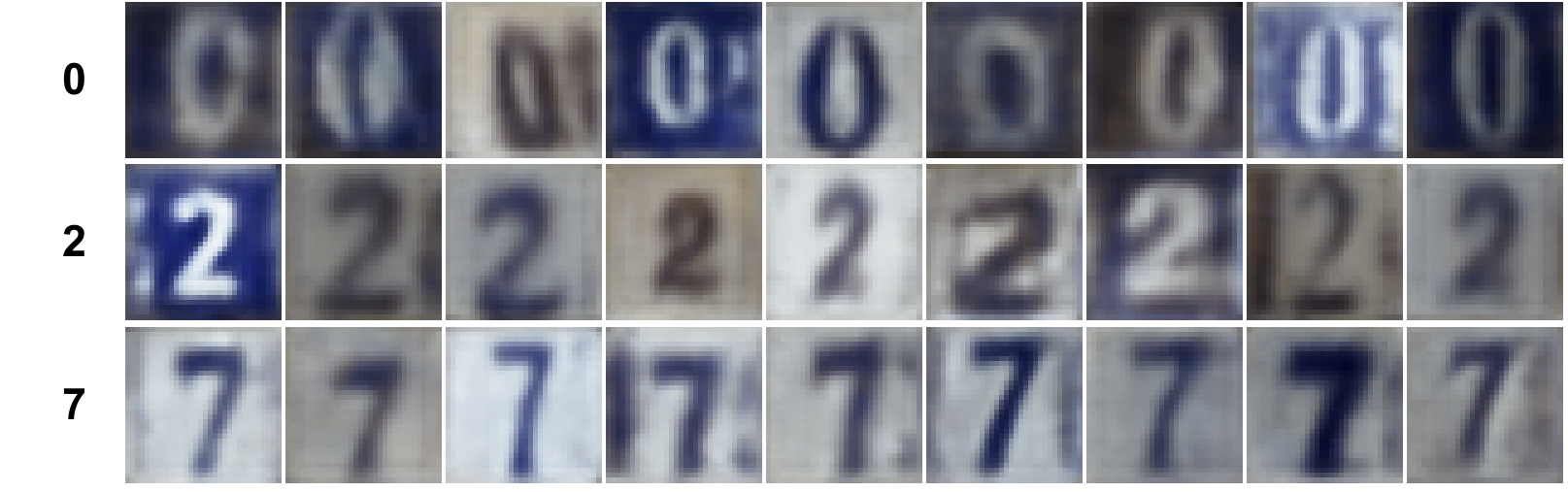}\vspace{-0.5em}
    \caption{SVHN}
    \includegraphics[width=1.0\textwidth]{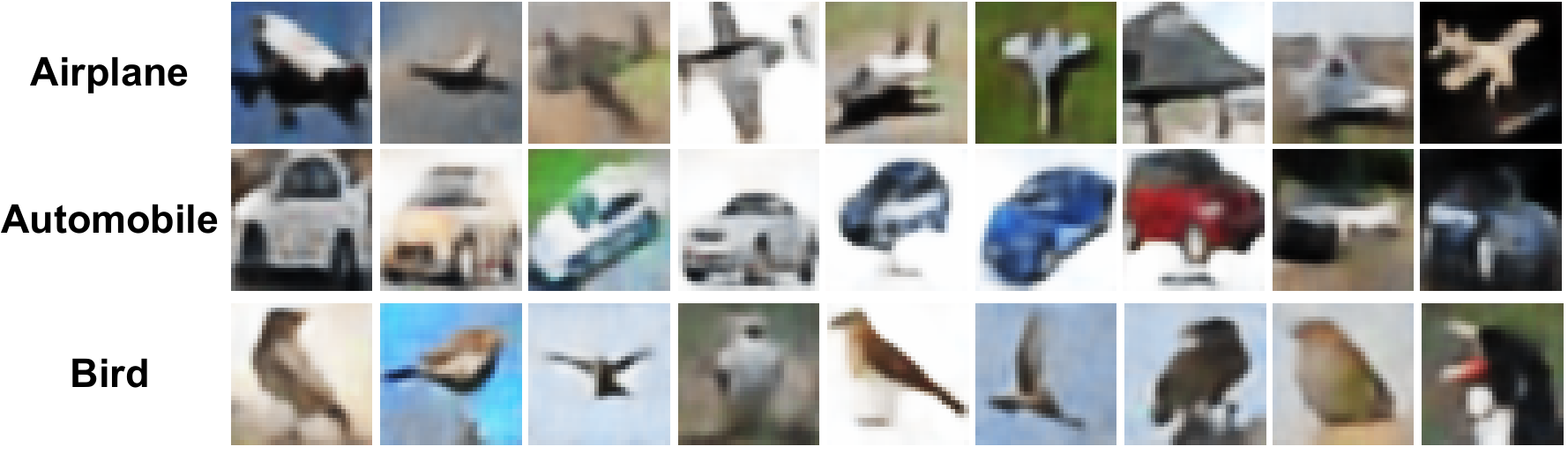}\vspace{-0.5em}
    \caption{CIFAR-10}
\end{subfigure}
\begin{subfigure}{0.51\columnwidth}
    \centering
    \includegraphics[width=1.0\textwidth]{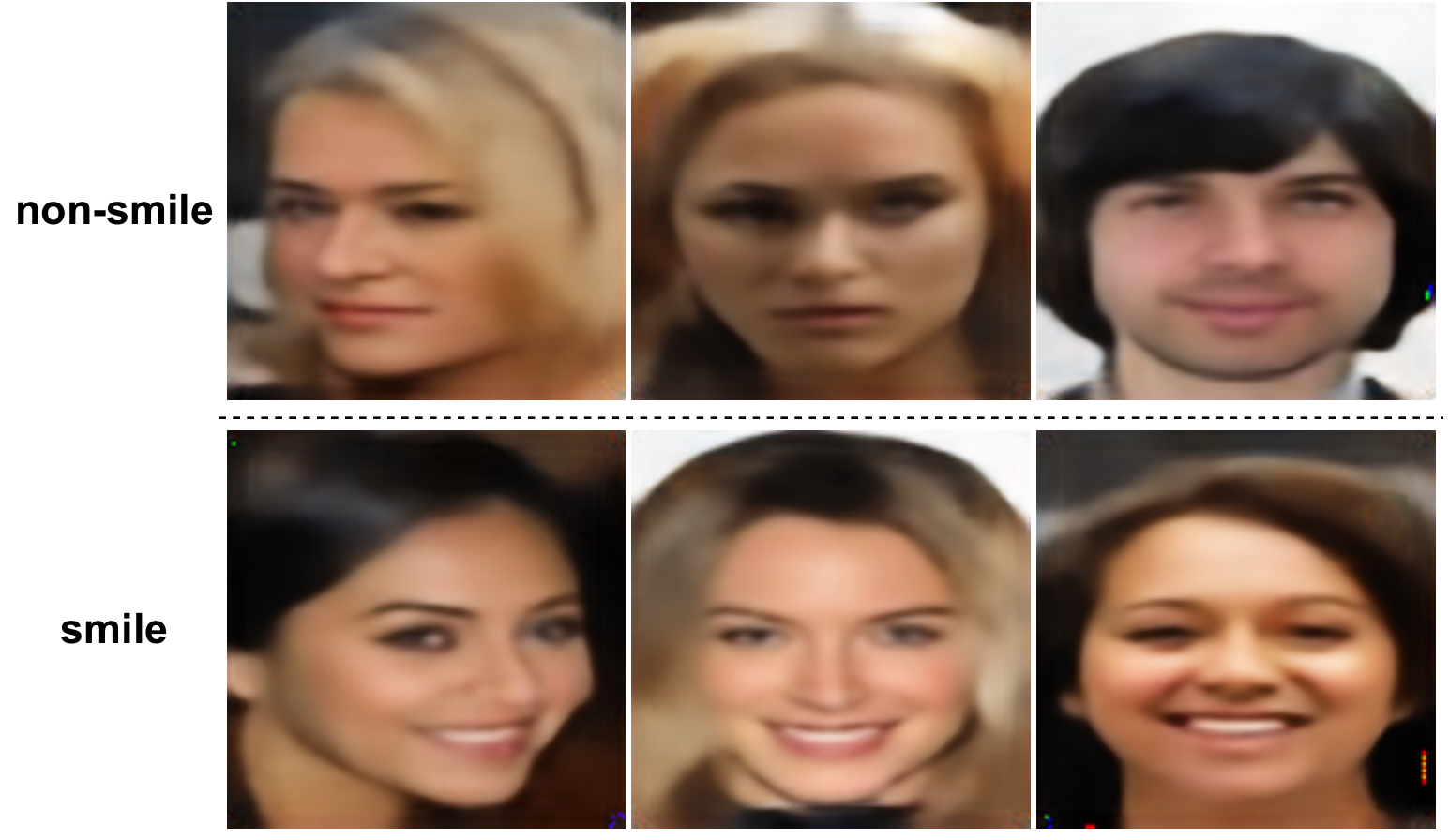}
    \caption{CelebA}
\end{subfigure}
    \caption{\rv{Adversarial examples generated by our LADDER method on SVHN, CIFAR-10 and CelebA, where the texts on the left indicate the actual class labels.}}
    \label{fig:advEx}
\end{figure*}

\rv{To demonstrate the quality of adversarial examples generated by LADDER on natural images, we show adversarial examples of selected classes on SVHN, CIFAR-10 and CelebA in Fig.~\ref{fig:advEx}. These examples are generated by perturbing the latent features with different perturbations $\epsilon$. As we can see, these generated images are of high quality without any pepper noise.} 

\subsubsection{Diversity of Generated Adversarial Examples}
\label{exp:adv_sample}

We compare adversarial examples generated by our LADDER method and \rv{other methods (FGSM, JSMA, PGD and \citep{song2018constructing})} on MNIST. For LADDER, we used the trained generator to generate adversarial examples against the vanilla LeNet. \rv{The latent features that input to the last fully connected layer} in LeNet are used to train a linear SVM which yields the normal of boundary for generation. Each extracted latent feature is changed by adding perturbations. Finally, perturbed latent features are fed into the trained generator to generate adversarial examples. 

\vspace{-0.5cm}
\begin{figure}[htbp]
    \centering
    \includegraphics[scale=1.1]{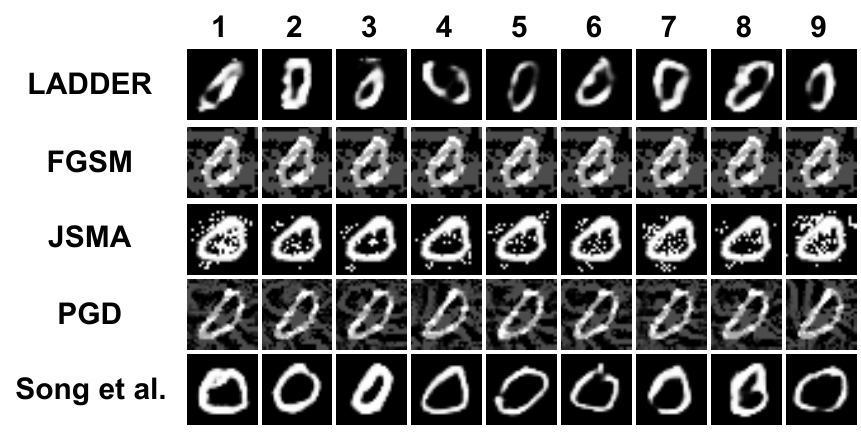}
    \caption{\rv{Adversarial examples generated by FGSM, JSMA, PGD, \citep{song2018constructing} and our LADDER method, where the topmost number indicates the predicted class label.}}
    \label{fig:our_AdSm}
\end{figure}

Fig.~\ref{fig:our_AdSm} shows  example images generated by FGSM, JSMA, \rv{PGD, \citep{song2018constructing}} and our LADDER method. Clearly, LADDER generates a more diverse set of distinct examples, \rv{whereas FGSM, JSMA and PGD tend to generate noisy images of repeating patterns.} This is because LADDER generates the examples by modifying latent features rather than slightly altering the original images in the input space. \rv{As compared with Song \etal~,  adversarial examples generated by LADDER are in general more visually diverse. This diversity property enables LADDER to be more effective for defending against adversarial attacks.}

\begin{figure}[tbp]
    \centering
    \includegraphics[width=0.85\textwidth]{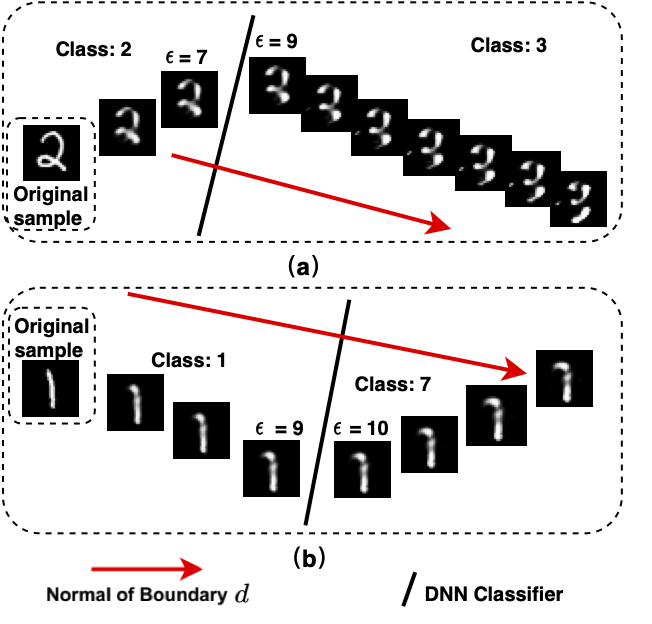}
    \caption{Generated adversarial examples by our LADDER method are high quality and they are generated near the decision boundary. The number on top of an image is the perturbation ($\epsilon$).}
    \label{fig:boudary}
\end{figure}

\subsubsection{High-Quality Adversarial Examples near Boundary}
\label{exp:quality}

Fig.~\ref{fig:boudary} shows adversarial examples generated by our LADDER method in relation to the decision boundary. Compared with adversarial examples generated by FGSM and JSMA (see Fig.~\ref{fig:our_AdSm}), LADDER is able to generate non-blurry images of adversarial examples that contain no noise in the background. Such high-quality adversarial examples would not hurt the standard accuracy.  

From classification perspectives, samples close to the decision boundary are more likely to be misclassified by a classifier. These samples should be more useful for constructing the classifier to obtain good standard accuracy. LADDER uses the normal of decision boundary as a guide to generate adversarial examples near the boundary. As shown in Fig.~\ref{fig:boudary}(a), the original sample is apparently a digit $2$. When we increase the perturbation ($\epsilon$) added to its corresponding latent features along the normal of the decision boundary, the predicted label of the generated samples changes from $2$ to $3$. The two examples near the decision boundary, generated with $\epsilon = 7$ and $\epsilon = 9$, are inherently ambiguous, even making humans difficult to make a judgement. If we add these ambiguous adversarial examples with labels to the training set, it would enrich the data space near the decision boundary, thereby improving the generalization of the trained classifier. This is the same case for Fig.~\ref{fig:boudary}(b). The DNN classifier predicts the two examples, generated with $\epsilon = 9$ and $\epsilon = 10$, as class 1 and class 7, while they look very similar. Such similar adversarial examples would be beneficial to improve the standard accuracy. {We provide more illustrative examples \rv{in Fig.~\ref{fig:geneMnist} in Appendix \ref{moreExp}} to demonstrate the ability of our generator to generate sensible adversarial examples and the effectiveness of  perturbing latent features along the normal of the decision boundary.}

\subsection{\rv{Standard Accuracy and Adversarial Robustness}}
\label{exp:cleanAndRobust}

\rv{To validate the efficacy of our LADDER method on standard accuracy, \ie, the accuracy on clean test datasets, as well as adversarial robustness, we conduct experiments on MNIST, SVHN, CelebA and CIFAR-10.} 

\rv{The results are reported in Tables 1--4, where row 1 indicates the vanilla model, and other rows indicate adversarially trained models; column 2 represents the clean test dataset, and columns 3--8 represent different attack methods that are used to generate adversarial examples for attacking the targeted model.} Under our setting, we focus on defence methods based on adversarial training, and each adversarially trained model is trained on adversarial examples generated by different methods under white-box setting. For FGSM and PGD, we set perturbation as 0.3; for CW, we choose $l_2$ norm distance. For CW and JSMA, the generation of adversarial examples is under targeted attack condition. For other methods, the generation is under untargeted attack condition. \rv{For LADDER, the architectures of the generators on four datasets are provided in Appendix~\ref{generatorArch}.}
\rv{To improve the generation performance on natural images, \ie, CelebA and CIFAR-10, we generalize LADDER by replacing the $L_p$ norm loss with an adversarial loss used in generative adversarial network (GAN) ~\citep{goodfellow2014generative} to train a stronger generator. This leads to a variant of our method called LADDER-GAN.}

In model sharing scenarios, after one trained model is released, it could be targeted by different attacks, \rv{which are unknown to the trained models. Thus, we focus on black-box attacks as indicated in columns 3--8, where adversarial examples are generated with no access to the trained models.} In particular, we assess the ability of each adversarially trained model to defend other types of attacks. Thus, the robustness results of each adversarially trained model are not reported against the same attack used to generate adversarial examples.

\rv{We focus on assessing the performance of different models in terms of both standard accuracy on clean test dataset and adversarial robustness. We thus calculate the average rank for each adversarially trained model to show its trade-off between standard accuracy and adversarial robustness against several other adversarial attacks. The average rank is calculated over the ranks of each adversarially trained model on clean test dataset and defending all other attacks, which is reported as the last column in the tables.} 

\subsubsection{\rv{Results on SVHN}}

\begin{table*}[tbp]
\centering
\caption{\rv{SVHN: Classification accuracy (\%) of vanilla and adversarially trained models on clean test dataset and adversarial examples. Smaller means better for the average rank (Avg. Rank). The best method is highlighted in bold and the second best is underlined.}}
\begin{tabular}{C|F|FFFF>{\color{black}\centering}m{0.7cm}>{\color{black}\centering}m{0.9cm}|>{\color{black}\centering\arraybackslash}m{0.5cm}}
\toprule
Defence Method    & Clean & FGSM & JSMA  & PGD   & CW    & Song \etal & Auto Attack & Avg. Rank \\
\midrule
Vanilla Model    & 93.85 & 25   & 34.04 & 17.16 & 86.78 & 99.42       & 54.62      & 4.00        \\
\midrule
FGSM Adv.         & 88.66 & -    & 37.49 & 20.62 & 83.07 & 97.51       & 66.36      & 4.83        \\
JSMA Adv.         & 91.04 & 28.4 & -     & 19.6  & 86.22 & 98.69       & 61.8       & \underline{3.83}        \\
PGD Adv.          & 87.75 & 34.2 & 42.18 & -     & 85.96 & 96.69       & 73.18      & 4.17        \\
CW Adv.           & 91.11 & 23.6 & 37.64 & 18.11 & -     & 98.16       & 63.42      & 4.33        \\
Song \etal~ Adv.  & 93.53 & 28   & 33.91 & 17.18 & 87.29 & -           & 56.27      & 4.00        \\
\midrule
LADDER\_cavRandom & 91.55 & 24.8 & 36.96 & 14.78 & 84.93 & 98.72       & 50.78      & 5.86        \\
LADDER\_Random    & 90.12 & 21   & 35.69 & 16.42 & 83.96 & 98.33       & 53.87      & 6.86        \\
LADDER            & 91.71 & 26.8 & 37.29 & 16.82 & 86.42 & 98.96       & 62         & \textbf{3.71} \\
\bottomrule
\end{tabular}
\label{tab:svhnattack2}
\end{table*}

\rv{Table~\ref{tab:svhnattack2}reports the standard accuracy on clean SVHN test dataset and adversarial robustness of different models against other adversarial attacks. We can see that, among all the adversarially trained models, LADDER achieves the second best standard accuracy (91.71\%) on clean test dataset, which lags behind the Song \etal~ Adv. model only. Compared with other models, LADDER achieves an improvement of $3.96\%$ and $3.05\%$ over PGD and FGSM, respectively. As compared with the other two variants, LADDER\_cavRandom and LADDER\_Random, LADDER performs better on clean test dataset.}

\rv{We also find that, LADDER achieves the best performance in terms of defending the \citep{song2018constructing} attack, compared with all other adversarially trained models. As for the overall performance on defending all attacks and on clean test dataset, LADDER achieves an average rank of 3.71, outperforming all other methods. This shows that LADDER achieves a better trade-off between standard accuracy and adversarial robustness. Compared with two variants LADDER\_cavRandom and LADDER\_Random, LADDER improves the average rank by 2.15 and 3.15, respectively. This validates the necessity of using the normal of decision boundary as guidance to generate adversarial examples.} 

\subsubsection{\rv{Results on MNIST}} 

\begin{table}[tbp]
\centering
\caption{\rv{MNIST: Classification accuracy (\%) of vanilla and adversarially trained LeNet on clean and adversarial examples. Smaller means better for the average rank (Avg. Rank). The best method is highlighted in bold and the second best is underlined.}}
\begin{tabular}{C|F|FFFF>{\color{black}\centering}m{0.7cm}>{\color{black}\centering}m{0.9cm}|>{\color{black}\centering\arraybackslash}m{0.5cm}}
\toprule
Defence Method    & Clean & FGSM & JSMA  & PGD   & CW    & Song \etal & Auto Attack & Avg. Rank \\
\midrule
Vanilla Model        & 99.13 & 46.6 & 93.91 & 29.93 & 99.09 & 99.82       & 99.56      & 3.57        \\
\midrule
FGSM Adv.         & 92.31 & -    & 83.67 & 80.91 & 90.58 & 95.53       & 92.22      & 6.50        \\
JSMA Adv.         & 98.56 & 57.8 & -     & 51.04 & 98.56 & 99.87       & 98.69      & 4.17        \\
PGD Adv.          & 90.79 & 76.2 & 83.31 & -     & 90.67 & 94.44       & 90.36      & 7.00        \\
CW Adv.           & 98.87 & 59.2 & 94.67 & 44.62 & -     & 99.91       & 99.36      & \textbf{3.17}        \\
Song \etal~ Adv.  & 97.23 & 55.6 & 89.16 & 49.91 & 96.87 & -           & 96.69      & 6.00        \\
\midrule
LADDER\_cavRandom & 99.01 & 54.8 & 92.76 & 48.2  & 98.56 & 99.78       & 98.89      & 5.00        \\
LADDER\_Random    & 98.99 & 64.4 & 92.93 & 56.87 & 98.33 & 99.89       & 99.04      & \underline{3.29}        \\
LADDER            & 99.12 & 55.8 & 93.13 & 49.2  & 98.9  & 99.82       & 99.36      & \underline{3.29}       \\
\bottomrule
\end{tabular}
\label{tab:mnistattack}
\end{table}

Table~\ref{tab:mnistattack} reports the classification results of the vanilla and adversarially trained LeNet models on the clean MNIST test dataset \rv{and adversarial robustness against other attacks. In terms of standard accuracy on clean test dataset, LADDER performs the best and LADDER\_cavRandom achieves the second best among all the adversarially trained models. The performance of LADDER ($99.12\%$) is very close to that of the vanilla model ($99.13\%$), with only $0.01\%$ difference.} Moreover, LADDER outperforms the baseline PGD Adv. by a large margin of $8.33\%$. LADDER is also observed to perform better than its two counterparts, LADDER\_cavRandom and LADDER\_Random, while LADDER\_cavRandom performs better than LADDER\_Random. 

\rv{In terms of adversarial robustness, it is clear to observe that LADDER improves the vanilla model in defending FGSM attack and PGD attack by $9.2$\% and $19.27$\%, respectively. When defending the JSMA attack, LADDER performs similarly to the vanilla model.} \rv{Among all attacks, LADDER achieves the best performance of defending the CW attack and AutoAttack, compared with other adversarially trained models.} \rv{Overall, LADDER and LADDER\_Random achieve an average rank of 3.29, which is the highest among all adversarially trained models except for CW Adv. model. Yet, LADDER achieves better performance than CW Adv. on clean test dataset and against PGD attack, and achieves the same performance against AutoAttack. LADDER\_cavRandom also outperforms FGSM, PGD and Song \etal~ Adv. model} This confirms the usefulness of leveraging the latent features to generate adversarial examples.

\subsubsection{\rv{Results on CelebA}}

\begin{table*}[tbp]
\centering
\caption{\rv{CelebA: Classification accuracy (\%) of vanilla and adversarially trained CelebANet on clean examples and adversarial examples. Smaller means better for average rank (avg.  Rank). The best method is highlighted in bold and the second best is underlined.}}
\begin{tabular}{C|F|FFFF>{\color{black}\centering}m{0.7cm}>{\color{black}\centering}m{0.9cm}|>{\color{black}\centering\arraybackslash}m{0.5cm}}
\toprule
Defence Method    & Clean & FGSM  & JSMA  & PGD   & CW    & Song \etal & Auto Attack & Avg. Rank \\
\midrule
Vanilla Model    & 91.4  & 52.65 & 83.05 & 13.9  & 62.1  & 92.05       & 49.30      & 5.43        \\
\midrule
FGSM Adv.         & 89.4  & -     & 67.15 & 18.55 & 62.95 & 54.95       & 53.95      & 7.67        \\
JSMA Adv.         & 90.45 & 53.1  & -     & 14.95 & 65.9  & 93.5        & 40.35      & 5.50        \\
PGD Adv.          & 89.55 & 51.75 & 63.65 & -     & 67.05 & 59.55       & 55.95      & 6.00        \\
CW Adv.           & 89.5  & 50.1  & 77.1  & 43.25 & -     & 77.8        & 68.35      & 5.67        \\
Song \etal~ Adv.  & 91.15 & 53.2  & 83.9  & 20    & 66.15 & -           & 49.95      & \underline{3.50}        \\
\midrule
LADDER\_cavRandom & 91.1  & 52.8  & 81.9  & 24.9  & 64.95 & 86.8        & 42.55      & 5.43        \\
LADDER\_Random    & 91.05 & 55.1  & 80.15 & 24.65 & 64.45 & 87          & 55.1       & 4.71        \\
LADDER-GAN        & 91.45 & 52.75 & 80.9  & 25.25 & 64.7  & 88          & 46.7       & 4.71        \\
LADDER            & 91.95 & 53.25 & 82.4  & 27.6  & 65.1  & 87.1        & 48.95      & \textbf{3.29}       \\
\bottomrule
\end{tabular}
\label{tab:celebAattack2}
\end{table*}

\rv{Table~\ref{tab:celebAattack2} reports standard accuracy and adversarial robustness of the vanilla model and different adversarially trained models on CelebA. In terms of standard accuracy on clean test dataset, LADDER yields the highest accuracy, while LADDER-GAN achieves the second best performance. For the two variants of LADDER, LADDER\_cavRandom performs better than LADDER\_Random, while both variants outperform the FGSM, PGD, JSMA and CW Adv. models. This shows that performing feature perturbations in the latent space is beneficial to achieve better standard accuracy.} 

\rv{As for the adversarial robustness performance against adversarial attacks, LADDER achieves better performance of defending FGSM, JSMA, PGD and \citep{song2018constructing} attacks, compared with most of the baseline models. In particular, for the PGD attack, LADDER improves the accuracy from $13.90\%$ to $27.60\%$.}
\rv{As a whole, LADDER achieves an average rank of 3.29, which is the best among all methods. The smaller the average rank, the better the overall performance of defending adversarial attacks and achieving standard accuracy simultaneously.} \rv{LADDER-GAN and LADDER\_Random both achieve an average rank of 4.71, which stands behind only Song \etal~ Adv. and LADDER. This proves the overall effectiveness of LADDER and its variants. }



\subsubsection{\rv{Results on CIFAR-10}}

\begin{table}[tbp]
\centering
\caption{\rv{CIFAR-10: Classification accuracy (\%) of vanilla and adversarially trained models on clean and adversarial examples. Smaller means better for the average rank (Avg. Rank). The best method is highlighted in bold and the second best is underlined.}}
\begin{tabular}{C|F|FFFF>{\color{black}\centering}m{0.7cm}>{\color{black}\centering}m{0.9cm}|>{\color{black}\centering\arraybackslash}m{0.5cm}}
\toprule
Defence Method   & Clean & FGSM  & JSMA  & PGD   & CW    & Song \etal & Auto Attack & Avg. Rank \\
\midrule
Vanilla Model   & 88.99 & 58.02 & 80.81 & 56.87 & 59.99 & 28.01       & 42.31      & 4.57        \\
\midrule
FGSM Adv.        & 67.21 & -     & 63.4  & 63.85 & 68.56 & 24.1        & 63.08      & 4.83        \\
JSMA Adv.        & 87.87 & 70.72 & -     & 73.54 & 78.76 & 29.61       & 67.63      & \textbf{2.17}        \\
PGD Adv.         & 79.12 & 70.71 & 75.7  & -     & 72.15 & 23.92       & 69.37      & 4.17        \\
CW Adv.          & 84.53 & 76.9  & 81.3  & 79.46 & -     & 28.15       & 77.55      & \textbf{2.17}        \\
Song \etal~ Adv. & 48.66 & 9.99  & 45.7  & 11.01 & 8.87  & -           & 10.75      & 7.33        \\
\midrule
LADDER-GAN       & 85.04 & 59.71 & 75.84 & 58.83 & 60.39 & 31.27       & 45.68      & \underline{3.86}        \\
LADDER           & 85.92 & 58.26 & 75.84 & 60.18 & 53.92 & 29.85       & 47.75      & 4.00   \\
\bottomrule
\end{tabular}
\label{tab:defenceCifar}
\end{table}

\rv{We also compare standard accuracy and adversarial robustness of LADDER with other baseline methods on CIFAR-10 -- a more challenging dataset for the generation task. Table~\ref{tab:defenceCifar} shows the classification results of the vanilla model and different adversarially trained models. We can see that, LADDER is the second best performer among all adversarially trained models, achieving an accuracy of 85.92\%. Only the JSMA Adv. model performs slightly better than LADDER with a small gap of 1.95\%. Compared to Song \etal, FGSM and PGD Adv., LADDER achieves significant improvements by 37.26\%, 18.71\%, and 6.8\%, respectively. The performance of LADDER-GAN slightly lags behind LADDER. This signifies the competitive performance of LADDER in achieving good standard accuracy on CIFAR-10.}


\rv{For the adversarial robustness, LADDER achieves the best performance when defending the \citep{song2018constructing} attack. In general, LADDER achieves a better average rank than FGSM, PGD and Song \etal~ based adversarially trained models. 
As the generation task on CIFAR-10 is more challenging, we also compare with LADDER-GAN. As can be seen, LADDER-GAN improves the average rank of LADDER from 4.0 to 3.86. Yet, we find that LADDER and LADDER-GAN perform worse than CW and JSMA adversarially trained models. This indicates that generator-based defence methods have difficulties in achieving the most appealing results on challenging datasets like CIFAR-10. Our findings reaffirm the results of \citep{song2018constructing} and those reported in~\citep{jang2019adversarial} where a recursive and stochastic generator is used to generate adversarial examples for adversarial training. We leave further investigation of this problem to future work.}


\subsubsection{Analyse of the Trade-off between Standard Accuracy and Robustness}
\label{exp:trade-off}

\begin{figure}[t]
    \center
    \includegraphics[width=0.8\textwidth]{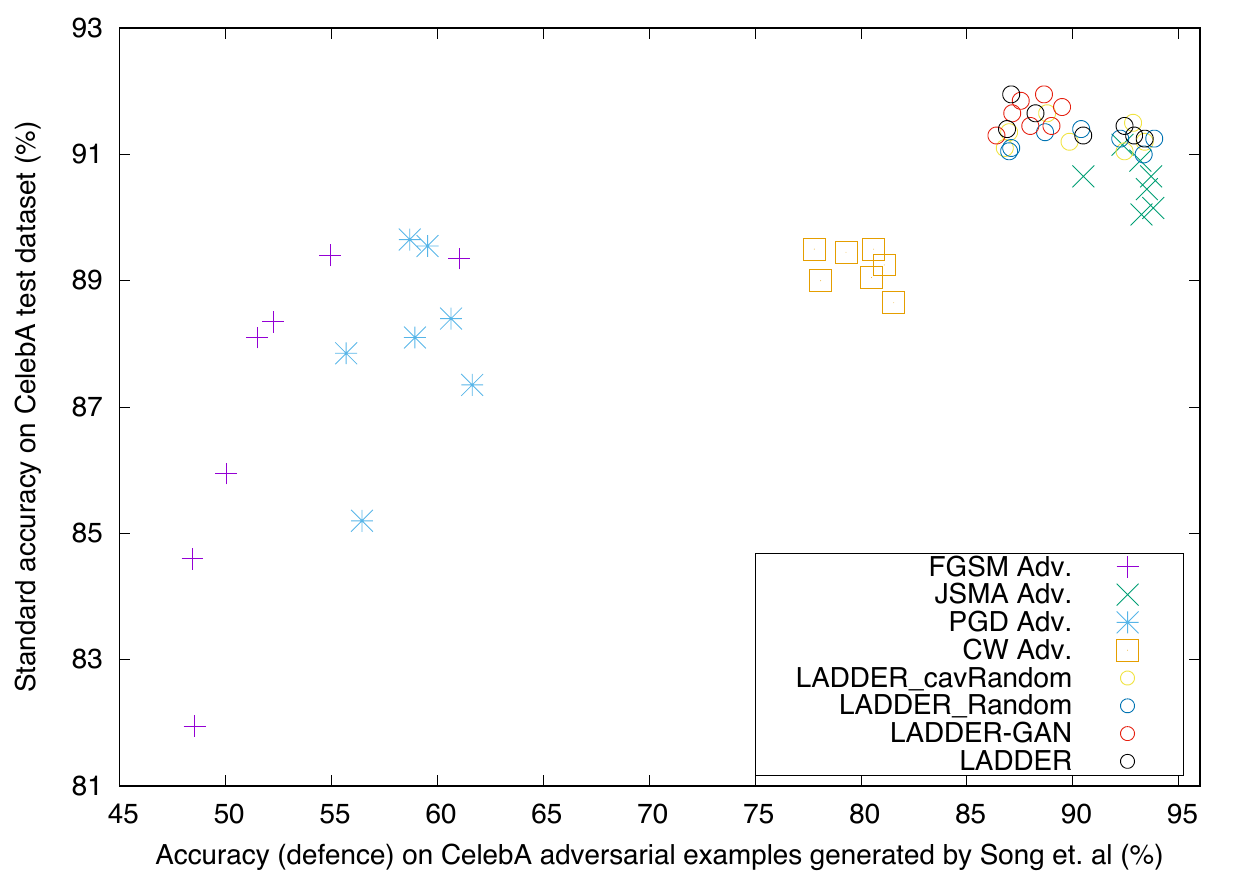}
    \caption{\rv{Classification accuracy of different defence methods on adversarial examples generated by Song \etal~\citeyearpar{song2018constructing} and on CelebA clean test dataset. Each model is adversarially trained on varying numbers of adversarial examples, with 7 points for each method compared in the figure. Best viewed in color.}}
    \label{fig:acc_trade-off}
\end{figure}

\rv{To visually demonstrate the advantages of our LADDER method in achieving a better trade-off between standard accuracy and adversarial robustness, we explicitly compare the trade-off performance of different defence methods with respect to different numbers of adversarial examples on CelebA as a case study. Specifically, we vary the number of examples used to adversarially train the models from 100 to 2,000. The classification results are plotted in Fig.~\ref{fig:acc_trade-off}, where there are 7 points for each adversarially trained model.} In the figure, the x-axis indicates the accuracy of adversarially trained models on adversarial examples generated by Song \etal~\citeyearpar{song2018constructing}, and the y-axis indicates the standard accuracy of adversarially trained models on clean CelebA test dataset. If the trade-off achieved by one method is better, the method is expected to locate in the top right corner. It can be seen clearly that, our LADDER method and its variants (marked in circles) are located in the top right corner. Markedly, our LADDER method outperforms FGSM Adv., PGD Adv., and CW Adv. by a large margin. Again, this confirms that our LADDER method is able to achieve a better trade-off between standard accuracy and adversarial robustness. 

\subsection{Effect of Perturbation} 
\label{exp:perturbation}

\begin{figure}[tbp]
    \center
    \includegraphics[width=0.85\textwidth]{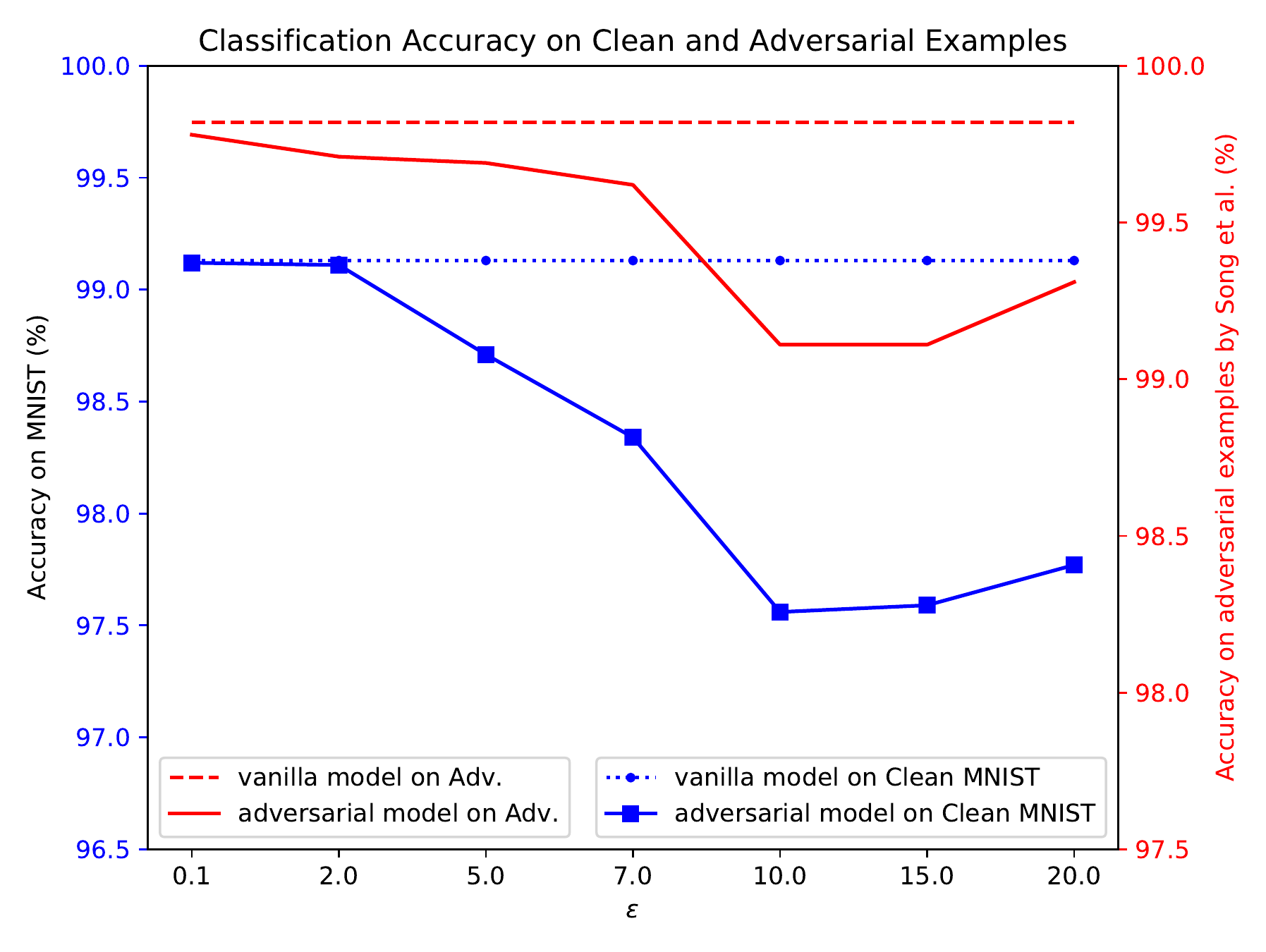}
    \caption{Classification accuracy of vanilla LeNet and adversarially trained LeNet on MNIST test dataset and \rv{adversarial examples with different perturbations} $\epsilon$. Best viewed in color.}
    \label{fig:acc_eps}
\end{figure}

Next, we empirically evaluate the effect of perturbation $\epsilon$ on the performance of our LADDER method. First, we study the impact of $\epsilon$ on standard accuracy. To adversarially train the LeNet, we randomly select 450 images per class from MNIST dataset to generate 4,500 adversarial examples for each perturbation [0.1, 2.0, 5.0, 7.0, 10.0, 15.0, 20.0]. These adversarial examples with different perturbations are separately used to adversarially train the LeNet. \rv{We undertake classification on clean MNIST test dataset using these adversarially trained LeNet models.} The results are reported in Fig.~\ref{fig:acc_eps}, colored in blue. It is clear to observe that: 1). As $\epsilon$ increases, the classification accuracy of the adversarially {trained} models decreases firstly and then slightly increases at a later stage. 2). \rv{With different $\epsilon$ values, the changes in classification accuracy are within an interval of 1.56\% only.} 3). When $\epsilon$ is not too large, \ie~ $<7$, the performance of the adversarially trained models and the vanilla LeNet is very close. 

Second, we study the impact of $\epsilon$ on adversarial robustness. We still use the adversarially trained LeNet models in the previous step for conducting experiments. These models are used to defend adversarial examples generated using the Song \etal~\citeyearpar{song2018constructing} attack method. From the red part in Fig.~\ref{fig:acc_eps}, we can see that, as $\epsilon$ increases, the performance of the adversarially trained model drops slightly. Overall, with different $\epsilon$ values, our LADDER method is able to achieve stable performance within a reasonably small range. 

\subsection{Complement to Regularization-based Adversarial Training Methods}
\label{exp:CompWithTrades}

\begin{table*}[tbp]
\centering
\caption{\rv{Classification accuracy (\%) of vanilla and adversarially trained models on clean test dataset and adversarial examples generated by different attack methods. Higher means better for classification accuracy. The best results are highlighted in bold.}}
\begin{tabular}{>{\color{black}\centering}m{1.13cm}|>{\color{black}\centering}m{2.55cm}|FFFFF>{\color{black}\centering}m{0.7cm}>{\color{black}\centering\arraybackslash}m{0.9cm}}
\toprule
Dataset  & Defence Method & Clean          & FGSM           & JSMA           & PGD            & CW             & Song \etal & Auto Attack     \\
\midrule                         
 \multirow{3}{*}{MNIST} 
                         & TRADES         & 98.44          & \textbf{60.8}  & 92.6           & \textbf{52.31} & 98.4           & 99.8           & 98.71          \\
                         & LADDER         & \textbf{99.12} & 55.8           & 93.13          & 49.2           & \textbf{98.9}  & \textbf{99.82} & 99.36          \\
                         & TRADES+LADDER  & 98.94          & 54.8           & \textbf{93.36} & 52.04          & 98.78          & \textbf{99.82} & \textbf{99.78} \\
\midrule                         
\multirow{3}{*}{SVHN}  
                         & TRADES         & 85.88          & 13             & 27.24          & 10.38          & 73.93          & 92.31          & 44.36          \\
                         & LADDER         & \textbf{91.71} & \textbf{26.8}  & 37.29          & 16.82          & \textbf{86.42} & \textbf{98.96} & 62             \\
                         & TRADES+LADDER  & 91.01          & 26.2           & \textbf{38.02} & \textbf{20.11} & 84.89          & 98.36          & \textbf{64.16} \\
\midrule
 \multirow{3}{*}{CelebA}                      
                         & TRADES         & \textbf{91.95} & 52.85          & 83.1           & 12.85          & 63.45          & 89.2           & \textbf{53.7}  \\
                         & LADDER         & \textbf{91.95} & \textbf{53.25} & 82.4           & \textbf{27.6}  & \textbf{65.1}  & 87.1           & 48.95          \\
                         & TRADES+LADDER  & 91.7           & 52.75          & \textbf{83.3}  & 18.05          & 64.55          & \textbf{93.25} & 50             \\
\midrule                         
\multirow{3}{*}{CIFAR-10} 
                         & TRADES         & 74.32          & 73.49          & 72.39          & 73.89          & 73.01          & 17.2           & 73.79          \\
                         & LADDER         & \textbf{85.92} & 58.26          & 75.84          & 60.18          & 53.92          & \textbf{29.85} & 47.75         \\
                         & TRADES+LADDER  & 80.7           & \textbf{75.83} & \textbf{76.58} & \textbf{77.04} & \textbf{76.20} & 28.18          & \textbf{76.78} \\
 \bottomrule
\end{tabular}
\label{tab:TRADES}
\end{table*}

Experiments are further performed to testify whether LADDER can complement the existing regularization-based adversarial training methods that regularize the adversarial loss to achieve a better trade-off between standard accuracy and adversarial robustness. TRADES~\citep{zhang2019theoretically} is a strong competing method in this category. To achieve the same objective, our LADDER method takes a complementary approach to generate better adversarial examples but to use the original adversarial training loss. We expect that the performance of TRADES could be improved in combination with LADDER. 

\rv{We perform experiments on MNIST, SVHN, CelebA and {CIFAR-10} to compare LADDER, TRADES and the combined TRADES+LADDER. The results are shown in Table~\ref{tab:TRADES}. As we can see, LADDER achieves better defence performance than TRADES in 18 out of 28 cases on the four datasets. Especially on SVHN, LADDER outperforms TRADES against all attacks and on clean test dataset. As expected, TRADES+LADDER is found to outperform TRADES in most cases (23 out of 28) on the four datasets. This proves that, by generating high-quality and diverse adversarial examples, LADDER can complement regularization-based methods that modify the adversarial training loss function to further improve the performance. }
\section{Conclusion and Future Work}
We proposed a novel adversarial training framework called \textit{Latent Boundary-guided Adversarial Training} (LADDER), which adversarially trains DNN models through adversarial examples generated based on decision boundary in the latent space. We analyzed that, LADDER can generate high-quality and diverse adversarial examples. After adversarial training on the generated adversarial examples, LADDER achieves a better trade-off between standard accuracy and  adversarial robustness. \rv{The effectiveness of our LADDER method was validated through extensive experiments on MNIST, SVHN, CelebA, and CIFAR-10. From the new angle of improving the generation of adversarial examples, we showed that our method is also able to complement the existing regularization-based adversarial training methods.} 

\rv{In the future, we will extend our work from the following aspects.} Firstly, \rv{our method generates adversarial examples by perturbing along the normal of decision boundary to reduce the level of minimal perturbations in the latent space. For inverting to the input space}, we will try to derive theoretical bounds about when the perturbations of our generated examples are narrower than the $L_p$ norm perturbations in the input space. Secondly, for complex datasets like CIFAR-10 and ImageNet, where the generation task is more challenging, we have made attempts to use an adversarial loss rather than the $L_p$ norm loss for training a strong generator. We will investigate \rv{how to generate better adversarial examples to boost the adversarial robustness on complex datasets. Finally, we would like to reduce the computational complexity of our proposed method by removing the generator and directly using the adversarial features vectors in the latent space for adversarial training. } 

\section{Appendix}

\subsection{\rv{Hyper-parameters in Experiments}}
\label{hyperpara}
\rv{The hyper-parameters used for adversarial training in our experimental part are summarized in the Table~\ref{tab:hyper}.}

\begin{table}[!h]
    \centering
    \begin{tabular}{C|C|C|C|C}
    \toprule
        Dataset & Learning Rate & Epochs & Batch Size & Optimizer (momentum; weight decay) \\
        \midrule
        MNIST & 0.01 & 100 & 64 & SGD (0.5; 0) \\
        SVHN & 0.001 & 100 & 128 & SGD (0.9; 5e-4) \\
        CelebA & 0.01 & 100 & 64 & SGD (0.5; 5e-4) \\
        CIFAR-10 & 0.001 & 100 & 128 & SGD (0.9; 5e-4) \\
    \bottomrule
    \end{tabular}
    \caption{\rv{Hyper-parameters of adversarial training for all methods.}}
    \label{tab:hyper}
\end{table}

\vspace{-1cm}
\subsection{{Robustness under White-box Attacks}}
\label{robustWhiteBox}
\subsubsection{{Robustness on Defending White-box Attacks}}

{To verify the robustness of our method in defending white-box attacks, we have conducted experiments under white-box settings, where attack methods generate adversarial examples with gradients available from the network. The comparison of our method and other baseline methods on MNIST are shown in Table~\ref{tab:white-boxattack}. As can be seen, our methods (LADDER and LADDER\_Random) are the best two performers for defending different types of attacks simultaneously. Especially, our LADDER method achieves the best defence performance, when defending CW attacks. Compared with the vanilla model, LADDER improves the performance against all attacks except for CW and \citep{song2018constructing}. Overall, our proposed method exhibits competitive performance in defending white-box attacks.}

\begin{table}[htbp]
\centering
\caption{{Defending white-box attacks targeted on LeNet: classification accuracy of the vanilla LeNet and adversarially trained LeNet models on white-box adversarial examples generated by different attack methods. Smaller means better for the average rank (Avg. Rank). The best method is in bold and the second best is underlined.}}
\begin{tabular}{c|ccccc|c}
\toprule
Defence Method & FGSM & JSMA & PGD & CW & Song \etal & Avg Rank \\
\midrule
Vanilla Model & 1.80\% & 90.72\% & 0.69\% & 96.88\% & 98.28\% & - \\
\midrule
FGSM Adv. & - & 82.04\% & 76.28\% & 86.88\% & 90.76\% & 5 \\
JSMA Adv. & 11.24\% & - & 2.08\% & 95.54\% & 98.65\% & 4.25 \\
PGD Adv. & 81.00\% & 80.39\% & - & 86.71\% & 92.40\% &  5.25 \\
CW Adv. & 6.29\% & 92.17\% & 1.41\% & - & 99.38\%  & 4 \\
Song \etal~ Adv. & 28.74\% & 85.63\% & 17.13\% &  93.83\% & - & 3.25 \\
\midrule
LADDER\_cavRandom & 15.31\% & 90.31\% & 4.60\% &  93.47\% & 94.19\% & 4.8 \\
LADDER\_Random & 25.80\% & 91.54\% & 8.52\% & 95.86\% & 98.51\% & \underline{3} \\
LADDER & 19.6\% & 92.07\% & 9.02\% & 96.70\% & 98.23\% & \textbf{2.8} \\
\bottomrule
\end{tabular}
\label{tab:white-boxattack}
\end{table}

\subsubsection{{LADDER's Robustness against LADDER Attacks}}
{We have also conducted experiments to compare the defence performance of the vanilla model and our trained model against white-box adversarial examples generated using LADDER. The results are reported in Table~\ref{tab:defenseLAD}. As can be clearly seen, our trained model (LADDER) significantly improves the defence performance of the vanilla model on the three datasets by 28.54\%, 31.56, and 5.95\%, respectively. This proves the efficacy of our trained models against the adversarial examples generated under white-box settings using the same latent based attack method. }

\begin{table}[htbp]
    \centering
    \caption{{Classification accuracy on white-box adversarial examples generated by LADDER.}}
    \begin{tabular}{c|cc}
    \toprule
    Dataset & Vanilla Model & LADDER \\
    \midrule
    MNIST & 69.78\% & \textbf{98.23\%} \\
    SVHN & 42.48\% & \textbf{74.04\%} \\
    CelebA & 83.85\% & \textbf{89.8\%} \\ 
    \bottomrule
    \end{tabular}
    \label{tab:defenseLAD}
\end{table}

\subsubsection{{Susceptibility of Baseline Methods Against LADDER Attacks}}
{In Table~\ref{tab:attackOthers}, rows 3-7 show the defence performance of five conventional adversarially trained models against the adversarial examples generated by LADDER on three datasets. As compared to the vanilla model, the best performer among the five conventional adversarially trained models improves the defence performance by only 3.55\% on MNIST. All five conventional adversarially trained models exhibit worse defence performance than the vanilla model on SVHN and CelebA. This confirms the susceptibility of the conventional adversarially trained methods to the adversarial examples generated using LADDER. In contrast, the adversarially trained LADDER model is very successful in defending against adversarial samples generated using LADDER.}
\begin{table}[htbp]
    \centering
    \caption{{Defence performance of the conventional adversarially trained models and our LADDER method against adversarial samples generated using LADDER. The best method is highlighted in bold.}}
    \begin{tabular}{c|cccc}
    \toprule
    Defence Method & MNIST & SVHN & CelebA \\
    \midrule
    Vanilla Model & 69.78\% & 42.48\% & 83.85\% \\
    \midrule
    FGSM Adv. & 66.81\% & 40.01\% & 73.00\% \\
    JSMA Adv. & 72.16\% & 39.68\% & 83.30\% \\
    PGD Adv.  & 67.65\% & 39.12\% & 75.35\% \\
    CW Adv. & 73.33\% & 42.23\% & 80.95\% \\ 
    Song \etal~ Adv. & 72.55\% & 42.16\% & 83.15\% \\
    \midrule
    LADDER & \textbf{98.23\%} & \textbf{74.04\%} & \textbf{89.8\%} \\
    \bottomrule
    \end{tabular}
    \label{tab:attackOthers}
\end{table}

\subsection{{Illustrative Examples Generated by LADDER}}
\label{moreExp}

{We further provide more illustrative examples to demonstrate the ability of our generator to generate sensible adversarial examples and the effectiveness of perturbing latent features along the normal of the decision boundary.}

\begin{figure}[htbp]
    \centering
    \includegraphics[width=1.0\textwidth]{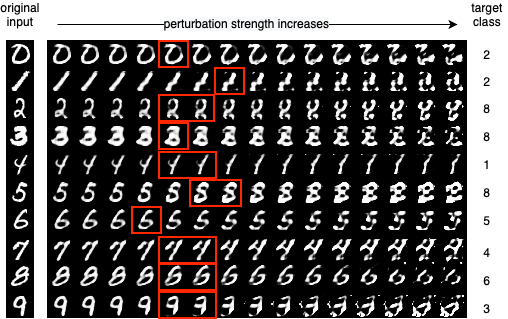}
    \caption{{Illustrative examples generated using our LADDER method with an increasing perturbation ($\epsilon$).}}
    \label{fig:geneMnist}
\end{figure}

{As shown in Fig.~\ref{fig:geneMnist}, the first column is the original input image; the last column is the randomly sampled target class; columns 2--16 are the generated examples, which are generated by adding different perturbations ($\epsilon$) to latent features of the original inputs. From column 2 to column 16, the perturbation increases gradually from 0.5 to 30.0 along the normal of decision boundary between the class of original input and target class. We can see from the figure, when we increase the perturbation, the generated examples gradually change from the original class to the target class, and when the perturbation is too large (\ie, the last 3 columns), the generated images are distorted. The images marked with red rectangles are inherently ambiguous between the class of the original input and the target class, even making humans difficult to make a judgement. These images enrich the data space near the decision boundary, thereby improving the generalization of the trained classifier.}





\subsection{{Generator Architectures}}
\label{generatorArch}
The neural network architectures of boundary-guided generator for MNIST, SVHN, CelebA and CIFAR-10 are detailed in this part. In each table, \textit{Linear} indicates linear transformation; \textit{Conv\_Transpose} denotes transposed convolution;  \textit{Conv} represents convolution; \textit{BN} represents batch normalization. \textit{kernels} means number of kernels. \textit{kernel} means the dimension of kernel. \textit{stride} means steps of convolutions. \textit{ReLU} means the ReLU activation function.

\begin{table}[htbp]
\caption{The architecture of boundary-guided generator for MNIST.}
\centering
\begin{tabular}{ccc}
\toprule
Layers & Layer parameters\\
\midrule
Linear & input: 500, output: 50 $\times$ 4 $\times$ 4 \\
Conv\_Transpose &  kernel: 2 $\times$ 2, stride: 4 $\times$ 4\\
Conv &  kernel: 3 $\times$ 3, stride: 1 $\times$ 1\\
Conv\_Transpose &  kernel: 2 $\times$ 2, stride: 3 $\times$ 3\\
Conv &  kernel: 4 $\times$ 4, stride: 1 $\times$ 1\\
Conv &  kernel: 5 $\times$ 5, stride: 1 $\times$ 1\\
\bottomrule
\end{tabular}
\label{app:mnist_g}
\vspace{-0.15cm}
\end{table}



\begin{table}[htbp]
\caption{The architecture of boundary-guided generator for SVHN.}
\centering
\begin{tabular}{ccc}
\toprule
Layers & Layer parameters\\
\midrule
Linear & input: 4096, output: 512 $\times$ 2 $\times$ 2 \\
Conv \& BN \& ReLU &  kernels: 512, kernel: 3 $\times$ 3, stride: 1\\
Conv\_Transpose \& BN & kernels: 512, kernel: 2 $\times$ 2, stride: 1\\
\midrule
Conv \& BN \& ReLU &  kernels: 512, kernel: 3 $\times$ 3, stride: 1\\
Conv\_Transpose \& BN & kernels: 512, kernel: 2 $\times$ 2, stride: 1\\
\midrule
Conv \& BN \& ReLU &  kernels: 256, kernel: 3 $\times$ 3, stride: 1\\
Conv\_Transpose \& BN & kernels: 256, kernel: 2 $\times$ 2, stride: 1\\
\midrule
Conv \& BN \& ReLU &  kernels: 128, kernel: 3 $\times$ 3, stride: 1\\
Conv\_Transpose \& BN & kernels: 128, kernel: 2 $\times$ 2, stride: 1\\
\midrule
Conv \& BN \& ReLU &  kernels: 64, kernel: 3 $\times$ 3, stride: 1\\
Conv \& Tanh &  kernels: 3, kernel: 1 $\times$ 1, stride: 1\\
\bottomrule
\end{tabular}
\label{app:svhngenerator}
\vspace{-0.1cm}
\end{table}


\begin{table}[!htbp]
\caption{The architecture of boundary-guided generator for CelebA.}
\centering
\begin{tabular}{cccc}
\toprule
Layers & Layer parameters & Repeat\\
\midrule
Linear & input: 4096, output: 512 $\times$ 4 $\times$ 4 & 1\\
Conv\_Transpose \& BN & kernels: 512, kernel: 2, stride: 1 & 1\\
Conv \& BN \& ReLU &  kernels: 512, kernel: 3, stride: 1 & 3\\
\midrule
Conv\_Transpose \& BN & kernels: 512, kernel: 2, stride: 1 & 1\\
Conv \& BN \& ReLU &  kernels: 512, kernel: 3, stride: 1 & 3\\
\midrule
Conv\_Transpose \& BN & kernels: 256, kernel: 2, stride: 1 & 1\\
Conv \& BN \& ReLU &  kernels: 256, kernel: 3, stride: 1 & 3\\
\midrule
Conv\_Transpose \& BN & kernels: 128, kernel: 2, stride: 1 & 1\\
Conv \& BN \& ReLU &  kernels: 128, kernel: 3, stride: 1 & 2\\
\midrule
Conv\_Transpose \& BN & kernels: 64, kernel: 2, stride: 1 & 1\\
Conv \& BN \& ReLU &  kernels: 64, kernel: 3, stride: 1 & 2\\
\midrule
Conv \& Tanh &  kernels: 3, kernel: 1, stride: 1 & 1\\
\bottomrule
\end{tabular}
\vspace{-0.2cm}
\label{app:celebAgenerator}
\end{table}

\begin{table}[!htbp]
\caption{{The architecture of boundary-guided generator for CIFAR-10.}}
\centering
\begin{tabular}{cccc}
\toprule
Layers & Layer parameters & Repeat\\
\midrule
Linear & input: 512, output: 512 $\times$ 4 $\times$ 4 & 1\\
Conv \& BN \& ReLU &  kernels: 512, kernel: 3, stride: 1 & 1\\
Conv\_Transpose \& BN & kernels: 512, kernel: 2, stride: 2 & 1\\
Conv \& BN \& ReLU &  kernels: 512, kernel: 3, stride: 1 & 4\\
\midrule
Conv\_Transpose \& BN & kernels: 512, kernel: 2, stride: 2 & 1\\
Conv \& BN \& ReLU &  kernels: 256, kernel: 3, stride: 1 & 4\\
\midrule
Conv\_Transpose \& BN & kernels: 256, kernel: 2, stride: 2 & 1\\
Conv \& BN \& ReLU &  kernels: 128, kernel: 3, stride: 1 & 4\\
\midrule
Conv \& BN \& ReLU &  kernels: 64, kernel: 3, stride: 1 & 2\\
Conv \& BN \& ReLU &  kernels: 32, kernel: 3, stride: 1 & 2\\
Conv \& BN \& ReLU &  kernels: 8, kernel: 3, stride: 1 & 2\\
\midrule
Conv \& Sigmoid &  kernels: 3, kernel: 1, stride: 1 & 1\\
\bottomrule
\end{tabular}
\vspace{-0.2cm}
\label{app:cifargenerator}
\end{table}


%
%
















\bibliographystyle{spbasic}      
\bibliography{mybibfile}   


\end{document}